\documentclass[journal]{IEEEtran}
\usepackage{times}
\usepackage{graphicx}
\usepackage{amsmath}
\usepackage{amssymb}
\usepackage{amsthm}
\usepackage{algorithm}
\usepackage{algorithmic}
\usepackage{multirow}
\usepackage{subfigure}
\usepackage{caption}
\usepackage{cases}
\usepackage{bm}
\usepackage{epsfig}
\usepackage{enumerate}
\usepackage{subfigure}
\usepackage{xcolor} 
\usepackage{etoolbox}
\usepackage{boxedminipage} 
\usepackage{booktabs} 
\DeclareMathOperator*{\argmin}{arg\,min}

\newcommand{\MyMapTemplatePrefix}[4]{\expandafter#1\csname#3#4\endcsname{#2{#4}}}
\newcommand{\MyMapTemplatePrefixNew}[5]{\expandafter#1\csname#4#5\endcsname{#2{#3{#5}}}}
\forcsvlist{\MyMapTemplatePrefix {\def} {\mathbf} {}} {A,B,C,D,E,F,G,H,I,J,K,L,M,N,O,P,Q,R,S,T,U,V,W,X,Y,Z}
\forcsvlist{\MyMapTemplatePrefix {\def} {\mathbf} {}} {a,b,c,d,e,f,g,h,i,j,k,l,m,n,o,p,q,r,s,t,u,v,w,x,y,z,1,0}
\forcsvlist{\MyMapTemplatePrefix {\def} {\widetilde} {wt}} {A,B,C,D,E,F,G,H,I,J,K,L,M,N,O,P,Q,R,S,T,U,V,W,X,Y,Z}
\forcsvlist{\MyMapTemplatePrefix {\def} {\widetilde} {wt}} {a,b,c,d,e,f,g,h,i,j,k,l,m,n,o,p,q,r,s,t,u,v,w,x,y,z} 
\forcsvlist{\MyMapTemplatePrefixNew {\def} {\widetilde}{\mathbf} {tb}} {A,B,C,D,E,F,G,H,I,J,K,L,M,N,O,P,Q,R,S,T,U,V,W,X,Y,Z}
\forcsvlist{\MyMapTemplatePrefixNew {\def} {\widetilde}{\mathbf} {tb}} {a,b,c,d,e,f,g,h,i,j,k,l,m,n,o,p,q,r,s,t,u,v,w,x,y,z}
\forcsvlist{\MyMapTemplatePrefix {\def} {\widehat} {wh}} {A,B,C,D,E,F,G,H,I,J,K,L,M,N,O,P,Q,R,S,T,U,V,W,X,Y,Z}
\forcsvlist{\MyMapTemplatePrefix {\def} {\widehat} {wh}} {a,b,c,d,e,f,g,h,i,j,k,l,m,n,o,p,q,r,s,t,u,v,w,x,y,z}
\forcsvlist{\MyMapTemplatePrefixNew {\def} {\widehat}{\mathbf} {hb}} {A,B,C,D,E,F,G,H,I,J,K,L,M,N,O,P,Q,R,S,T,U,V,W,X,Y,Z}
\forcsvlist{\MyMapTemplatePrefixNew {\def} {\widehat}{\mathbf} {hb}} {a,b,c,d,e,f,g,h,i,j,k,l,m,n,o,p,q,r,s,t,u,v,w,x,y,z}
\forcsvlist{\MyMapTemplatePrefix {\def} {\mathcal}{mc}} {A,B,C,D,E,F,G,H,I,J,K,L,M,N,O,P,Q,R,S,T,U,V,W,X,Y,Z}
\forcsvlist{\MyMapTemplatePrefix {\def} {\mathbb} {mb}} {A,B,C,D,E,F,G,H,I,J,K,L,M,N,O,P,Q,R,S,T,U,V,W,X,Y,Z}
\forcsvlist{\MyMapTemplatePrefix {\DeclareMathOperator} {} {} } {tr,diag,sgn}
\def\tp{^\intercal} 
\def\ie{\emph{i.e.}} \def\etal{\emph{et al.}}
 \def\eg{\emph{e.g.}}

\graphicspath{}

\def\iccvPaperID{2463} 

\begin{document}

\title{Spectral Filter Tracking}
\author{Zhen~Cui$^*$$^\dag$,~\IEEEmembership{Member,~IEEE,}
        Youyi~Cai$^\dag$,
        Wenming~Zheng,~\IEEEmembership{Member,~IEEE,}
        Jian~Yang,~\IEEEmembership{Member,~IEEE,}

\thanks{Zhen Cui is with the School of Computer Science and Engineering, Nanjing University of Science and Technology, Nanjing, China.\protect\\ 
E-mail: zhen.cui@njust.edu.cn,\protect\\}

\thanks{Youyi Cai and Wenming Zheng are with the Key Laboratory of Child Development and Learning Science of Ministry of Education, Research Center for Learning Science,
Southeast University, Nanjing, Jiangsu 210096, China.\protect\\ 
E-mail: yy\_cai@seu.edu.cn,wenming\_zheng@seu.edu.cn,.}

\thanks{Jian Yang is with the School of Computer Science and Engineering, Nanjing University of Science and Technology, Nanjing, China.\protect\\
E-mail:csjyang@njust.edu.cn.
Asterisk indicates corresponding author.\protect\\}
}

\maketitle

\begin{abstract}
Visual object tracking is a challenging computer vision task with numerous real-world applications. Here we propose a simple but efficient Spectral Filter Tracking (SFT) method. To characterize rotational and translation invariance of tracking targets, the candidate image region is models as a pixelwise grid graph. Instead of the conventional graph matching, we convert the tracking into a plain least square regression problem to estimate the best center coordinate of the target. But different from the holistic regression of correlation filter based methods, SFT can operate on localized surrounding regions of each pixel (i.e., vertex) by using spectral graph filters, which thus is more robust to resist local variations and cluttered background. To bypass the eigenvalue decomposition problem of the graph Laplacian matrix $\mcL$, we parameterize spectral graph filters as the polynomial of $\mcL$ by spectral graph theory, in which $\mcL^k$ exactly encodes a k-hop local neighborhood of each vertex. Finally, the filter parameters (i.e., polynomial coefficients) as well as feature projecting functions are jointly integrated into the regression model.

SFT can simply boil down to only a few line codes, but surprisingly it beats the correlation filter based model with the same feature input, and achieves the current best performance on the dataset~\cite{wu2015object} under the same feature extraction strategy (i.e., the existing VGG-Net model~\cite{simonyan2014very}). The code will be fully released in our website soon.
\end{abstract}

\section{Introduction}

Visual object tracking is a fundamental task in computer vision, due to its wide applications to video surveillance, traffic monitoring, and augmented reality, etc. Despite significant advance that has been achieved in visual tracking over the past few decades, this task remains very challenging because of unpredictable appearance variations including partial occlusion, geometric deformation, illumination change, background clutter, fast motion, etc.

The typical visual tracking starts with an initial bounding box of an object at the first frame, and then sequentially predicts the locations of the target in the next frames. To attain robust tracking, numerbers of tracking methods have sprung up. Among the existing tracking works, those part-based methods~\cite{adam2006robust,cehovin2013robust,liu2015real} have drawn increasing attention due to their robustness to local occlusion or appearance variations. They usually partition the object target (or the candidate region) into several parts and extract some useful cues from these parts. In the part-based methods, the topology structures (\eg, tree or graph)~\cite{kwon2013highly,cai2013structured} are often used to characterize the relationship of parts, and then some voting or matching strategies \cite{adam2006robust,zhang2014partial} are taken to find those reliable parts. In principle the part-based model is robust to resist partial occlusions and local appearance variations, but in practice it is difficult for accurate part partition, even though several methods~\cite{adam2006robust,kwon2009tracking} have been developed. More recently the tracking-by-segmentation methods~\cite{ren2007tracking,wang2011superpixel,hong2014tracking,wen2015jots} attempt to accurately annotate foreground and background regions based on the superpixel techniques. But the segmentation is usually time-consuming, and its results heavily influence on the tracking performance.

In contrast, the holistic tracking methods are more popular, especially the recent correlation filter (CF) based model. Due to  high-efficiency and excellent robustness, the correlation filter (CF) based model has aroused wide attention ~\cite{bolme2010visual,danelljan2014accurate,danelljan2014adaptive,li2014scale,ma2015long,henriques2015high} in the field of visual tracking. CF based methods attempt to learn a group of discriminative correlation filters that can produce correlation peaks for the tracking targets while suppressing their responses on background regions. To speed up the tracker, the original convolutional filters are transformed into frequency domain, and then learnt by scanning the candidate regions on a circular sliding window. As a holistic model, the CF based methods identically treat the whole candidate region, so those cluttered background might affect the trackers and degrade the tracking performance. To address this problem, some regularization methods of correlation filters~\cite{danelljan2015learning,cui2016recurrently} are proposed to spatially suppress background regions. But the holistic CF based methods is flexible enough not to resist local appearance variations like those part-based methods.


In this paper, we propose a simple but efficient Spectral Filter Tracking (SFT) method. To model rotational and translation invariance of the tracking targets, we construct a pixel-level grid graph for a candidate image region, which thus avoids any operations of part partition or superpixel segmentation. Moreover, the vertexes of the graph enclose the center of the tracking target, so we only need discover the best matching vertex from this graph. But instead of the conventional graph matching, we convert it into a plain least square regression problem to estimate the best center coordinate of the target.

But different from the holistic regression of CF based methods, we regress the tracking model on multiple localized regions for each pixel (\ie, vertex). To extract the local regions associated with each vertex, we use those spectral filters of graph. To solve those spectral filters, we need to decompose the graph Laplacian matrix. To bypass the eigenvalue decomposition problem, the spectral filters are parameterized as the polynomial of graph Laplacian matrix, in which the $k$-th term of Laplacian matrix exactly defines a $k$-localized spatial region. Consequently spectral filtering on graph is approximately equal to the operating on graph Laplacian matrix. By using the terms of the Laplacian polynomial as the filter bases, we jointly learn the parameters (\ie, polynomial coefficients) as well as feature projection by feeding responses of filter bases into the regression model.

Finally, the proposed tracker SFT can simply boil down to only a few line codes, but surprisingly the experimental results on the dataset~\cite{wu2015object} show that, the SFT beats the CF based model under the condition of the same input feature, surpasses the recent CF methods~\cite{danelljan2015convolutional,qi2016hedged} on the localization accuracy, and achieves the current best performance under the same feature extraction strategy (\ie, the existing VGG-Net model~\cite{simonyan2014very}).

\section{Related Work}

Video object tracking has been extensively studied over the past decades. They usually fall into two categories: generative model~\cite{babenko2009visual,liu2011robust,zhong2012robust,ross2008incremental} and discriminative model~\cite{avidan2004support,hare2016struck,henriques2015high}. Generative methods search for the best matching regions of the tracked target. Discriminative methods learn a classification model to distinguish the target from the backgrounds. Below we briefly introduce the main related work, including the correlation filter based methods and the part based methods.

Recently the correlation filter based discriminative model has aroused wide attention in the field of visual tracking. After Minimum Output Sum of Squared Error (MOSSE)~\cite{bolme2010visual} filter was proposed, numerous correlation filter methods start flowing in the field of computer vision~\cite{chen2015experimental}. Henriques \etal~\cite{henriques2012exploiting} used the kernel trick, and Danelljan \etal~\cite{danelljan2014adaptive} represented the inputs with color attributes. To handle the target scale problem, SAMF~\cite{li2014scale}, DSST~\cite{danelljan2014accurate} and an improved KCF~\cite{henriques2015high} were proposed subsequtially and achieved state-of-the-art performance. With more related methods developed~\cite{ma2015long,liu2015real,li2015reliable}, correlation filter based trackers have demonstrated their robustness. Especially, by employing high-level convolutional features as the inputs, correlation filter based methods~\cite{danelljan2015convolutional,qi2016hedged} achieved the current best performance on the public visual tracking dataset~\cite{wu2015object}. As a holistic filtering model, correlation filter based methods easily absorb those clutter information of background region while making full use of the background information. To reduce the influence of clutter background, some regularization methods~\cite{danelljan2015learning,cui2016recurrently} were proposed to suppress the response of background region by weighting  correlation filters. Although the regularization strategy has demonstrated the initial success on suppressing the background, it is still lack of an intrinsic revision for the holistic model. Different from these correlation filter based methods, we directly perform locally filtering on pixel-level graph structure.

In contrast to correlation filter based methods, the part-based methods~\cite{adam2006robust,ross2008incremental,cehovin2013robust,liu2015real} seems be more silent recently.
They usually partition the object target into several parts, and then attempt to discover some useful cues from the reliable parts. Adam \etal~\cite{adam2006robust} partitioned the object into several fragments, and then employed the voting strategy to decide the target position. Jia \etal~\cite{jia2012visual} selected the closest candidate patches of the next frame by using $l_1$ sparsity. Kwon \etal~\cite{kwon2009tracking} modeled local patches in topology structure in order to find reliable parts. Zhang \etal~\cite{zhang2014partial} performed part matching among multiple frames. Liu \etal~\cite{liu2015real} learned one response function on each part, and integrated all response maps to decide the final tracking confidence. Besides, the topology structure is used to model the relationship of the parts, \eg, tree structure~\cite{kwon2013highly} or graph structure on superpixels~\cite{cai2013structured}. In principle the part-based model is a robust solution to resist partial occlusion and local appearance variations. But in practice it is difficult to accurately define the partition of local parts, even though several strategies~\cite{adam2006robust,kwon2009tracking} have been developed. To address this problem, the tracking-by-segmentation methods~\cite{ren2007tracking,wang2011superpixel,hong2014tracking,wen2015jots} attempt to accurately annotate foreground and background regions by using the superpixel segmentation technique. But the tracking performance heavily depends on the segmentation results, and the superpixel segmentation is rather time-consuming. Different from these part-based methods, our proposed method performs local spectral filtering on the pixel-grid graph structure and then convert the target localization as a simple regression problem.

\section{Spectral Filter Tracker}\label{sec: sft}

\subsection{Overview}\label{sec: overview}

\begin{figure*}[t]
\centering
\includegraphics[width=0.8\linewidth]{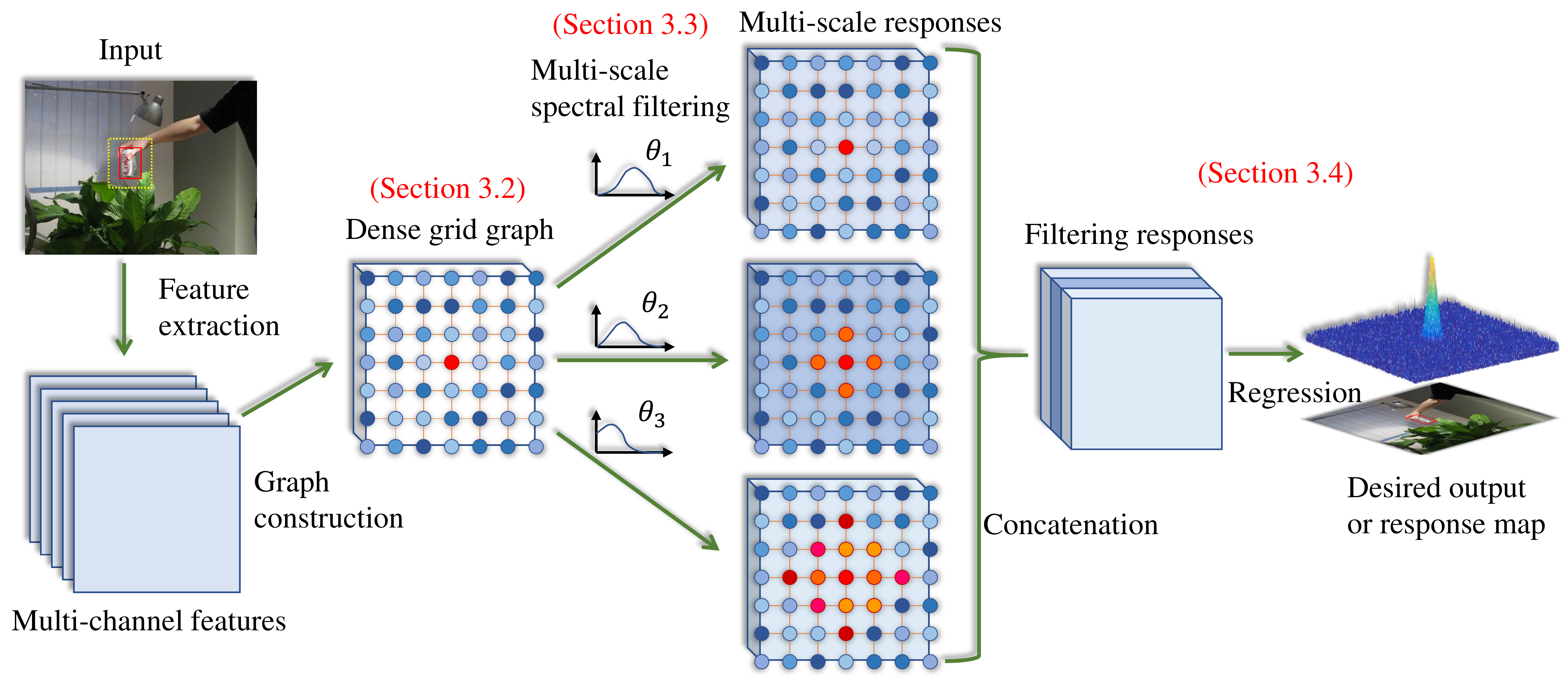}
\caption{The flowchart of spectral filter tracking. More details can be found in Section~\ref{sec: overview}.
}
\label{fig:framework}
\end{figure*}

An overview on the SFT flowchart is shown in Fig.~\ref{fig:framework}. Given a video frame, we first determine a small candidate region  around the bounding box localized from the previous frame, considering the motions of targets in continuous video frames are usually subtle. To enhance the discriminability, we can represent the candidate region with hand-crafted descriptors (\eg, HOG~\cite{felzenszwalb2010object}) or convolutional features~\cite{simonyan2014very}. Thereafter we can obtain multi-channel features, where each spatial pixel position is associated with a multi-channel feature vector. To reduce the effect of local appearance variations in the tracking, we model the candidate region as a pixelwise grid graph (Section \ref{sec: graph}), which has rotation-invariant and shifting-invariant property. In the graph,one spatial pixel is regarded as one vertex of the graph, and the edges connect those spatial adjacent vertexes. This problem becomes the conventional graph matching. But generally the solution of graph matching is rather complex, which might involve the integer programming.

To bypass graph matching, we use spectral graph theory to analyze graph structure. Instead of the holistic filtering in CF based trackers, we perform local filtering on the graph structure (Section \ref{sec: cons_lsf}). By using spectral graph filters, we can derive out the responses on localized graph regions for each vertex. But it involves eigenvalue decomposition of graph Laplacian matrix. To avoid this operation, we parameterize spectral graph filters as a polynomial of Laplacian matrix. Each entry of the polynomial actually plays the role of localized filtering on graph. It means that the polynomial terms enclose different scale spectral graph filters. By using the polynomial terms as the basic filters, we can obtain the corresponding multi-scale features for each vertex, which well-model local information of graph. For each vertex, by concatenating its multi-scale responses to form the final representation, finally we feed the final representation into the regression model to jointly solving those filter parameters (\ie, polynomial coefficients) and feature projecting functions (Section~\ref{sec: regression}).

\subsection{The Representation of Graph}\label{sec: graph}

We model the pixelwise spatial grid structure as an undirected weighted graph. The weighted graph $\mcG=\{\mcV,\mcE,\W\}$ consists of a set of vertices $\mcV (|\mcV|=N)$ and a set of edges $\mcE$. The adjacency matrix $\W$ assigns positive values to those connected edges. Besides, each vertex is associated with a signal, \ie, here a multi-channel feature vector extracted from its coordinate position. Formally, the feature extraction function $f:\mcV\rightarrow \mbR^d$ defines the signals of vertexes, where $d$ is the feature dimension.

In spectral graph theory, a crucial operator is the graph Laplacian operator $\mcL$. The operator is defined as $\mcL = \D-\W$, where $\D\in\mbR^{N\times N}$ is the diagonal degree matrix with $D_{ii}=\sum_{j}W_{ij}$. An popular option is the normalized graph Laplacian, where each weight $W_{ij}$ is multiplied by a factor $\frac{1}{\sqrt{D_{ii}D_{jj}}}$, \ie,
\begin{align}
\mcL^{norm} = \D^{-\frac{1}{2}}\mcL\D^{-\frac{1}{2}}= \I-\D^{\frac{1}{2}}\W\D^{\frac{1}{2}},
\end{align}
where $\I$ is the identity matrix. Unless otherwise specified, the Laplacian matrix used below is the normalized version.

As a real symmetric matrix, the graph Laplacian $\mcL$ has a complete set of orthonormal eigenvectors. The eigenvectors $\{\u_l\}$ satisfy $\mcL\u_l = \lambda_l\u_l$ for $l=1,2,\cdots,N$, where $\{\lambda_l\}$ are nonnegative real eigenvalues. We assume all eigenvalues are ordered as $0=\lambda_1<\lambda_2\leq\lambda_3\cdots\leq\lambda_N=\lambda_{max}$. In matrix expression, the Laplacian matrix is decomposed into $\mcL=\U\Lambda\U\tp$, where $\Lambda=\diag([\lambda_1,\lambda_2,\cdots,\lambda_N])$.  Analogous to the classic Fourier transform, the graph Fourier transform of a signal $\x$ in spatial domain can be defined as $\hbx = \U\tp\x$~\cite{shuman2013emerging},
where $\hbx$ is the produced frequency signal. The corresponding inverse Fourier transform is
$\x = \U\hbx$.

\subsection{The Construction of Local Spectral Filters}\label{sec: cons_lsf}

Suppose $g(\cdot)$ is a filter function of the graph $\mcL$, we can define the frequency filtering on the input signal $\x$ as
$\whz(\lambda_l)=\whx(\lambda_l)\whg(\lambda_l)$, or the inverse graph Fourier transform,
\begin{align}
z(i) = \sum_{l=1}^{N} \whx(\lambda_l)\whg(\lambda_l)\whu_l(i),
\end{align}
where $\whz(\lambda_l), \whx(\lambda_l), \whg(\lambda_l)$ are the Fourier coefficients corresponding to the spectrum $\lambda_l$.
By using matrix notation, the signal $\x$ is filtered as
\begin{align}
\z = \whg(\mcL)\x =\U\left[\begin{array}{ccc}\whg(\lambda_1) & \cdots & 0 \\ \vdots & \ddots & \vdots\\ 0 & \cdots &\whg(\lambda_{N})\end{array}\right]\U\tp\x.\label{eqn:5}
\end{align}
Given the input $\x$ and the output $\z$, we need to solve the filter function $g(\cdot)$ in Eqn.~(\ref{eqn:5}), which requires eigenvalue decomposition. To reduce computation cost, a low order polynomial may be used to approximate $\whg(\cdot)$ in the frequency domain. Here we use the Chebyshev expansion of $K$ order~\cite{hammond2011wavelets}, which is defined by the recurrent relation $T_k(x)=2xT_{k-1}(x)-T_{k-2}(x)$ with $T_0=1$ and $T_1=x$. In the appropriate Sobolev space, the set of Chebyshev polynomials form an orthonormal basis, so any one function in the space $x\in[-1,1]$ may be expressed via the expansion: $f(x)=\sum_{k=0}^\infty a_kT_k(x)$.

To make the eigenvalues $\{\lambda_l\}$ of the Laplacian matrix $\mcL$ fall in $[-1,1]$, we may scale and shift them as $\widetilde{\lambda_l}=\frac{2}{\lambda_{max}}\lambda_l-1$, and then employ the Chebyshev polynomials on $\{\widetilde{\lambda_l}\}$.
If we consider a linear combination of the polynomial components, the $K$-order filter can be written as,
\begin{align}
\whg(\lambda_l) = \sum_{k=0}^{K-1} \theta_{k}T_k(\widetilde{\lambda_l}), \label{eqn:6}
\end{align}
where $\theta\in\mbR^K$ is a parameter vector of the polynomial coefficients, and $K$ is the order of the polynomial. By putting Eqn.~(\ref{eqn:6}) into Eqn.~(\ref{eqn:5}), we can have
\begin{align}
\z &= \U\left[\begin{array}{ccc}\sum_{k=0}^{K-1} \theta_{k}T_k(\widetilde{\lambda_l}) & \cdots & 0 \\ \vdots & \ddots & \vdots\\ 0 & \cdots &\sum_{k=0}^{K-1} \theta_{k}T_k(\widetilde{\lambda_l})\end{array}\right]\U\tp\x\nonumber\\
& = \sum_{k=0}^{K-1}\theta_k \U\left[\begin{array}{ccc}T_k(\widetilde{\lambda_l}) & \cdots & 0 \\ \vdots & \ddots & \vdots\\ 0 & \cdots &T_k(\widetilde{\lambda_l})\end{array}\right]\U\tp\x \label{eqn:7}\\
& = \sum_{k=0}^{K-1}\theta_k T_k(\U\left[\begin{array}{ccc}\widetilde{\lambda_l} & \cdots & 0 \\ \vdots & \ddots & \vdots\\ 0 & \cdots &\widetilde{\lambda_l}\end{array}\right]\U\tp)\x \label{eqn:8}\\
& = \sum_{k=0}^{K-1}\theta_k T_k(\frac{2}{\lambda_{max}}\mcL-\I)\x.\label{eqn:9}
\end{align}
From Eqn.~(\ref{eqn:8}) to Eqn.~(\ref{eqn:9}), we use the spectral decomposition of Laplacian matrix,   $\mcL=\U\diag([\lambda_1,\cdots,\lambda_N])\U\tp$. From Eqn.~(\ref{eqn:7}) to Eqn.~(\ref{eqn:8}), we utilize the basic calculation on the filter function, \ie,
\begin{align}
\mcL^k&=\U\diag([\lambda_1^k,\cdots,\lambda_N^k])\U\tp\nonumber\\
&=(\U\diag([\lambda_1,\cdots,\lambda_N])\U\tp)^k.
\end{align}

According to graph theory, $\mcL^k$ encodes a k-hop local neighborhood of each vertex. Consequently, the $K$-order polynomial in Eqn.~(\ref{eqn:9}) is a exactly $K$-localized filter function on the Laplacian graph. To obtain the local filtering responses on graph, thus we only need operate the Laplacian matrix $\mcL$. That means, each entry of the polynomial can be regarded as the filter bases, and $\theta$ is the parameters to be solved.

\subsection{The Prediction of the Tracking Target}\label{sec: regression}

In the visual tracking, we need predict the centers of the tracking target. Similar to those CF based methods, we regress a peak map $\y\in \mbR^{N\times 1}$ from the multi-channel features $\X\in \mbR^{N\times d}$, where $N$ is the number of pixels (\ie, vertexes) within the candidate region, each row of $\X$ corresponds to a signal of one vertex. Now we denote $\widetilde{\mcL}=\frac{2}{\lambda_{max}}\mcL-\I$. Then the filter bases defined in the polynomial of Eqn.~(\ref{eqn:9}) become $\{T_0(\widetilde{\mcL}), T_1(\widetilde{\mcL}), \cdots, T_{K-1}(\widetilde{\mcL})\}$, where the $k$-th filter basis only relates to the k-hop neighbor vertexes. Given a signal $\x$ and the filter parameter $\theta=[\theta_0,\theta_2,\cdots,\theta_{K-1}]$, we can obtain the local filtering response $\z$ if employing a linear combination of $K$ filter bases. We use the $k$ filter bases to filtering the graph, and then combine the learning of filter parameters and feature projecting function into a least square regression model,
\begin{align}
\underset{\w}{\argmin} \quad \|\mcF(\X)\w-\y\|^2 + \gamma\|\w\|^2, \label{eqn:11}
\end{align}
where $\gamma$ is the balance parameter, and $\mcF(\X)$ concatenates the responses of $K$ filter bases in feature dimensionality,
\begin{align}
\mcF(\X)=[T_0(\widetilde{\mcL})\X, T_1(\widetilde{\mcL})\X, \cdots, T_{K-1}(\widetilde{\mcL})\X]. \label{eqn:11b}
\end{align}
Thus the tracking model can be easily solved as
\begin{align}
\w = (\mcF(\X)\tp\mcF(\X))^{-1}\mcF(\X)\tp\y. \label{eqn:12}
\end{align}

\subsection{The Algorithm}\label{sec: alg}

We summarize the whole tracking algorithm in Alg.~\ref{alg:SFT}. There are two crucial steps, including the computation of the filtering responses and the regressor. As the spectral filter bases can be pre-computed before detecting targets, the computation cost mainly spends on the matrix inverse operation in Eqn.~{(\ref{eqn:12})}. The computation complexity is about $O(d^3K^3)$, where $d$ is the feature dimension and $K$ is the filter order. To speed up SFT, we can project features into low-dimension space by using Principal Components Analysis (PCA) or random projecting (Section \ref{sec: pca}). Besides, the order $K$ may be downscaled by designing the skipping neighborship (Section \ref{sec: graphcons}). For other strategies of accelerating matrix inverse calculation, we leave them as the future work.

\begin{algorithm}[t]
\caption{Spectral Filter Tracking Algorithm}\label{alg:SFT}
\begin{algorithmic}[1]
\REQUIRE A video sequence with initial target position at the fist frame; The regression map $\y$ (Gaussian-shape).
\ENSURE The coordinates of the tracking target.
\STATE Initialize the Laplacian matrix $\mcL$.
\REPEAT
  \IF{$t>1$}
     \STATE Extract feature $\X$ from the candidate region.
     \STATE Compute $K$-order responses $\mcF(\X)$ in Eqn.~(\ref{eqn:11b}).
     \STATE Compute detection score $\tby=\mcF(\X)\w$ Eqn.~(\ref{eqn:12}).
     \STATE Find the target center with maximum score.
  \ENDIF
\STATE Extract feature $\X$ at the current target location.
\STATE Compute $K$-order responses $\mcF(\X)$ in Eqn.~(\ref{eqn:11b}).
\STATE Derive the regression model $\w_t$ in Eqn.~(\ref{eqn:12}).
  \STATE\textbf{if} ($t=1$) \textbf{then} $\w=\w_t$;
  \STATE\textbf{else} $\w=(1-\alpha)*\w+\alpha*\w_t$.
\UNTIL{All frames are traversed.}
\end{algorithmic}
\end{algorithm}

\section{Implemental Details}

In this section, we introduce more details of SFT, including how to construct graph, how to reduce feature dimensions, and how to process the scale problem, etc.

\subsection{Graph Construction} \label{sec: graphcons}

\begin{figure}[t]
\centering
\includegraphics[width=1\linewidth]{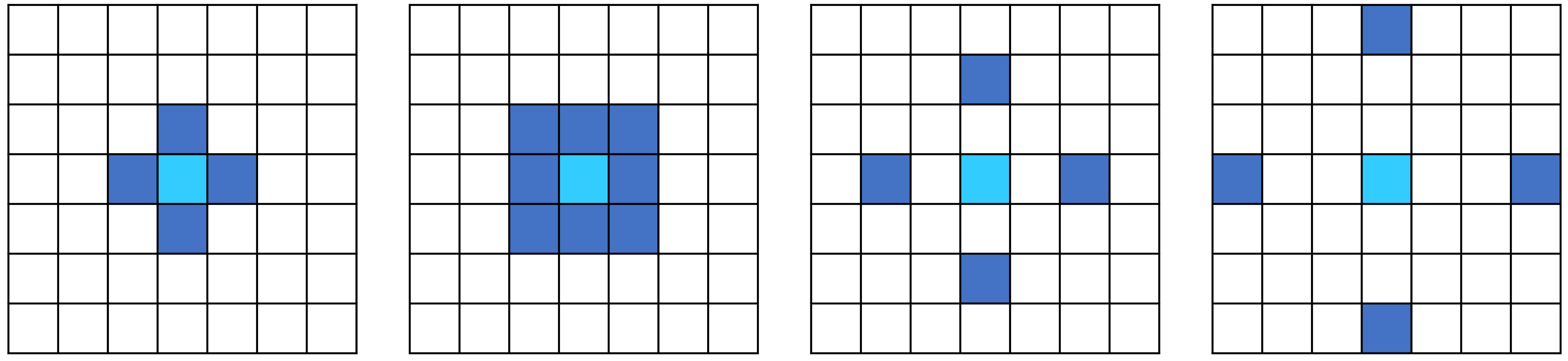}
\caption{Some exemplars of adjacency relationship. As image textures of adjacent pixels are high-degree similar, we can skip several pixels to connect edges as shown in the right two cases. The advantage is that the filter order $K$ can be decreased in proportion for the same size filtering region.
}
\label{fig:neighbors}
\end{figure}

As shown in Fig.~\ref{fig:neighbors}, we define the neighborship based on spatial layout. As analyzed in Section \ref{sec: cons_lsf}, the filter basis $T_k(\widetilde{\mcL})$ is an exactly $k$-localized filter, where the neighborship is propagated in $\mcL^k$. Thus we only need to define the spatial nearest neighbors of each reference point as shown in the first two cases of Fig.~\ref{fig:neighbors}, and employ the $k$ order filters to evolve neighborship. Considering the high-degree similarity of image textures in adjacent pixels, we may skip several pixels to connect edges as shown in the last two cases of Fig.~\ref{fig:neighbors}. Thus, when filtering on the same size region, the skipping mode need less filters (\ie, a smaller $K$). Consequently, the tracker speeds up if the smaller $K$ is employed, because the complexity of matrix inverse is related to $K$.

After choosing neighbourship, we may assign Gaussian weights or $\{0,1\}$ weights to those connected edges. To simplify the weighting step, here we use the $\{0,1\}$ weighting strategy, \ie, the adjacency matrix $\W$ of the weighted graph $\mcG$ is defined as
\begin{eqnarray}
W_{ij} = \left\{
         \begin{array}{cc}1, & \text{if $e\in \mcE$ connects vertices $i$ and $j$},\\
                          0, & \text{otherwise}.
         \end{array}
         \right.
\end{eqnarray}

\subsection{Multi-channel Features} \label{sec: pca}

The deep VGG-Net~\cite{simonyan2014very} is used to extract high-level features. We crop an image patch with 2.4 times the size of target bounding box and then resize it to $224\times 224$ pixels for the VGG-Net with 19 layers. Similar to the literature~\cite{qi2016hedged}, we use the outputs from six convolutional layers ($\{10,11,12,14,15,16\}$-th layers) as six types of feature maps. All feature maps are resize to $57\times 57$ pixel size, which is about quadruple to the smallest size ($14\times 14$) of feature maps. The odd size can uniquely define the target center. As the number of feature maps in each layer is 512, SFT will spend high computation cost on the matrix inverse as analyzed in Section \ref{sec: alg}, if all feature maps are concatenated together ($512\times6=3072$ dimensions). To speed up the tracking, two strategies are taken for the convolutional features: (i) We learn six trackers respectively corresponding to six layer features, and average the six tracking target centers as the final target center. (ii) We employ PCA to project each layer features into 100 dimensions.

\subsection{Other Details}

For the scale estimation, we employ the same strategy to ~\cite{danelljan2014accurate}. The filters at multiple resolutions are used to estimated scale changes in the target size. We extract the samples with sizes in scaled factors $a^r (r\in\{\lfloor\frac{1-S}{2}\rfloor,\cdots,\lfloor\frac{S-1}{2}\rfloor\})$ at the previous target location. The scales $a^r$ is relative to the current target scale, and $S$ is the number of scales and $a$ is the scale increment factor. In our experiment, we follow the same parameter settings to ~\cite{danelljan2014accurate}, where $S=33, a=1.02$.

For the filter order $K$, we make the largest filter basis (\ie, $T_{K-1}(\widetilde{\mcL})$) cover the whole target region. Suppose the target size is $h\times w$, the filter order $K$ is set to $\max(h,w)$ if choosing the nearest spatial neighborship (\eg, the first two cases in Fig.~\ref{fig:neighbors}). For the skipping modes in the last two cases of Fig.~\ref{fig:neighbors}, $K$ is assigned to $\lceil\frac{\max(h,w)}{s}\rceil$, where $s$ is the skipping step.

The default neighbor relationship uses the third case in Fig.~\ref{fig:neighbors}. The balance parameter $\gamma$ is set to 1.

\section{Experiment}

\subsection{Dataset and Setting}

In order to verify our proposed tracker, we conduct extensive experiments on OTB-2015 dataset~\cite{wu2015object}. The dataset consists of 100 video sequences with 11 different attributes including illumination changes, scale variation, motion blur, fast motion, etc. Two widely used evaluation criteria, \ie, precision plot and success plot, are used in the following experiments.


Precision plot measures center location error (CLE) which computes the difference between prediction positions and ground truth. It shows how many percentage of frames whose center location error is within a previously given threshold. Here the threshold score is set to 20 pixels. Success plot denotes bounding box overlap ratio which is based on area under the curve (AUC). The overlap ratio is defined as $S = |r_{t} \cap r_{0}|/|r_{t} \cup r_{0}|$, where $\cup$ and $\cap$ are the union and intersection operators, $r_{0}$ is the predicted bounding box and $r_{t}$ is the ground truth . More details can be found in~\cite{wu2015object}.


\subsection{Selection of Adjacent Vertexes}

As discussed in Section~\ref{sec: graphcons}, we only need to connect those nearest neighbors, as spectral graph filters can propagate the neighbor relationship to distant vertexes. Here we test four cases of Fig.~\ref{fig:neighbors}, whose results are reported in Tab.~\ref{tab:neighbor}. From this table, we can observe that, (i) more neighbors (Case 2) slightly degrade the performance, which may be attribute to feature confusion after averaging features of all neighbors during compute $\mcL^k\X$; (ii) the skipping mode with one pixel interval (Case 3) reaches the best performance. The skipping strategy can be regarded as the downsampling for feature maps. Thus the increase of skipping step (Case 4) degrades the performance because some useful information can not be encoded in the filtering process. Thus, we use the third case as the default setting in the following experiments.
\begin{table}[!hbt]
\setlength{\tabcolsep}{4pt}
\centering
\caption{The performance of different neighborhood strategies as shown in Fig.~\ref{fig:neighbors}. Case 1$\sim$4 are respectively correspond to the sequential subfigures. \textbf{Note that here we don't perform scale estimation}.}
\begin{tabular}{c|c|c|c|c|c|c|c}
\hline
\multicolumn{2}{c|}{Case 1} &
\multicolumn{2}{c|}{Case 2} &
\multicolumn{2}{c|}{Case 3} &
\multicolumn{2}{c}{Case 4} \\ \hline
CLE & AUC & CLE & AUC & CLE & AUC & CLE & AUC \\ \hline
0.855 & 0.572 & 0.847 & 0.565 &  0.866 & 0.576 & 0.862 & 0.574 \\ \hline
\end{tabular}
\label{tab:neighbor}
\end{table}

\subsection{Comparisons with CF Based Trackers}

\begin{figure}[h!]
\centering
\subfigure[]{\label{fig:5-1}
\includegraphics[width=0.47\linewidth]{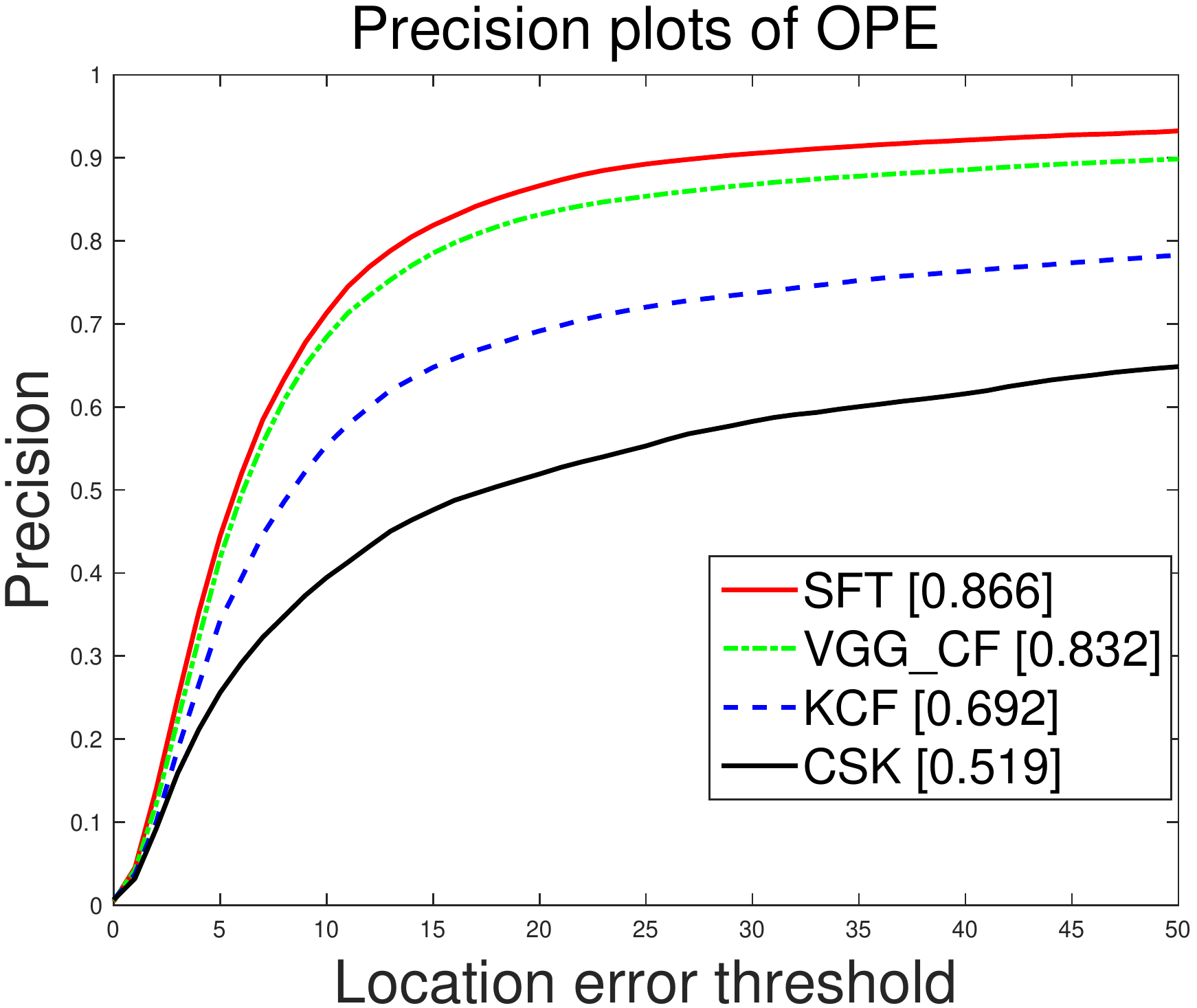}}
\subfigure[]{\label{fig:5-2}
\includegraphics[width=0.47\linewidth]{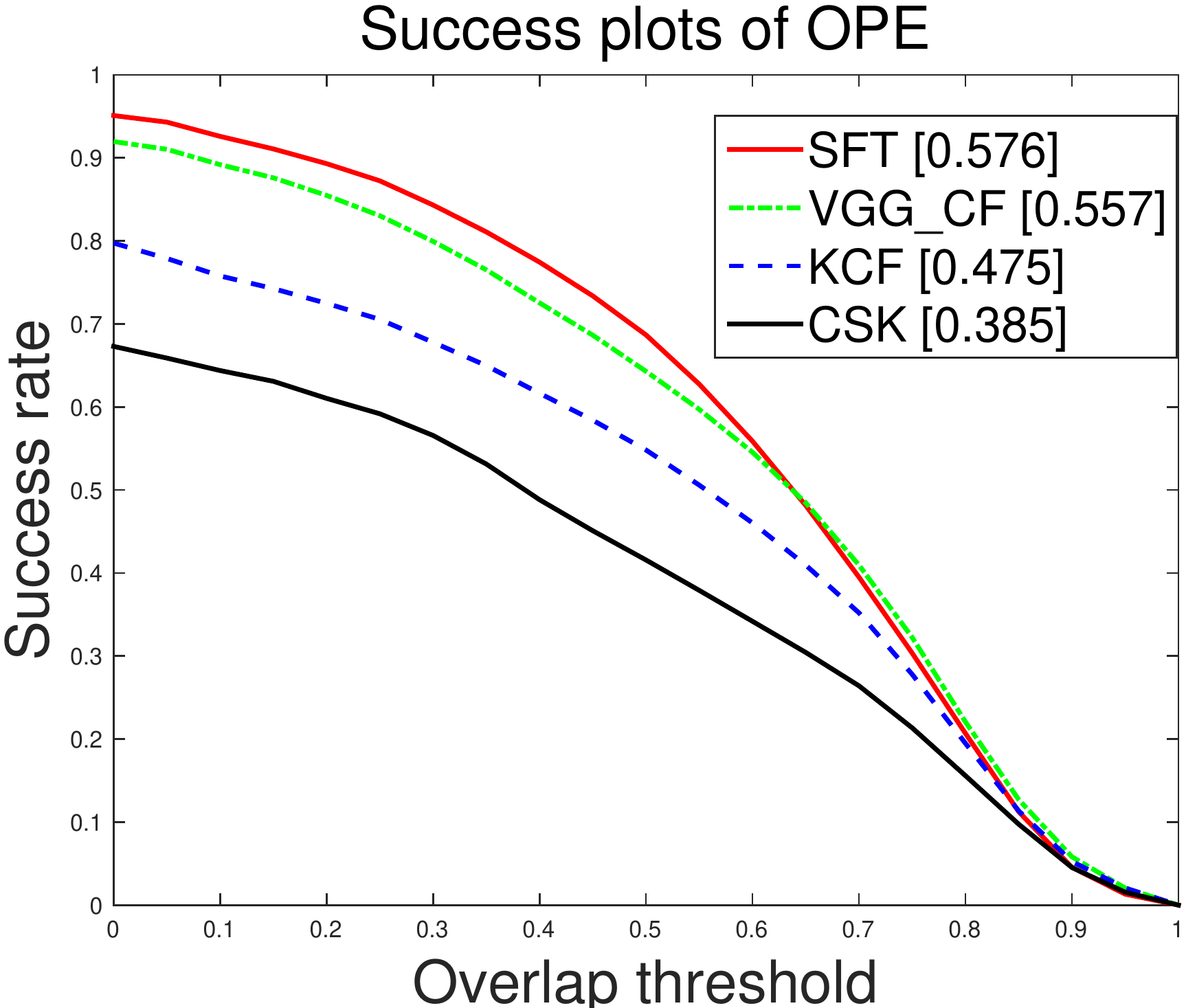} }
\caption{The precision and success plot of comparisons with CF based methods. \textbf{Note that here we don't perform scale estimation for an intrinsic comparison}.}
\label{fig:compare-correlation}
\end{figure}

To fairly compare the CF based model, the same features of VGG-Net are feed into the CF based model, which is our standard baseline, called VGG\_CF. Besides, we compare the classic CF based methods, CSK~\cite{henriques2012exploiting} and KCF~\cite{henriques2015high}.
Fig.~\ref{fig:compare-correlation} shows the results under the precision plots of One-Pass Evaluation(OPE) and the success plots based on area under curve(AUC). The performances of the three CF based methods are quite different. As the CSK trackers only use raw feature and CSK use HOG feature while the robust deep CNN feature is employed for VGG\_CF tracker, which makes it outperform the other two CF based methods obviously. Compared to the baseline VGG\_CF, our proposed SFT achieves a gain of 3.4\% in CLE. Meanwhile we obtains an AUC scores of 57.6\% which also outperforms VGG\_CF tracker. The reason may be two folds: (i) local filtering on spatial regions, (ii) rotation-invariance and shifting-invariance for graph structure. To implement an intrinsic comparison, here we don't process the scale.

\subsection{Comparisons with State-of-the-art}

We compare our proposed SFT with nine state-of-the-art trackers: DeepSRDCF~\cite{danelljan2015convolutional}, MEEM~\cite{zhang2014meem}, HDT~\cite{qi2016hedged}, CSK~\cite{henriques2012exploiting}, KCF~\cite{henriques2015high}, DSST~\cite{danelljan2014accurate}, SCM~\cite{zhong2012robust}, STRUCK~\cite{hare2016struck}, TLD~\cite{kalal2010pn}. The DeepSRDCF and HDT are two recent representative deep learning based trackers. Others employ HOG feature mostly. Only the top-10 trackers are reported in the experiments.

\begin{figure}[h!]
\centering
\subfigure[]{\label{fig:6-1}
\includegraphics[width=0.47\linewidth]{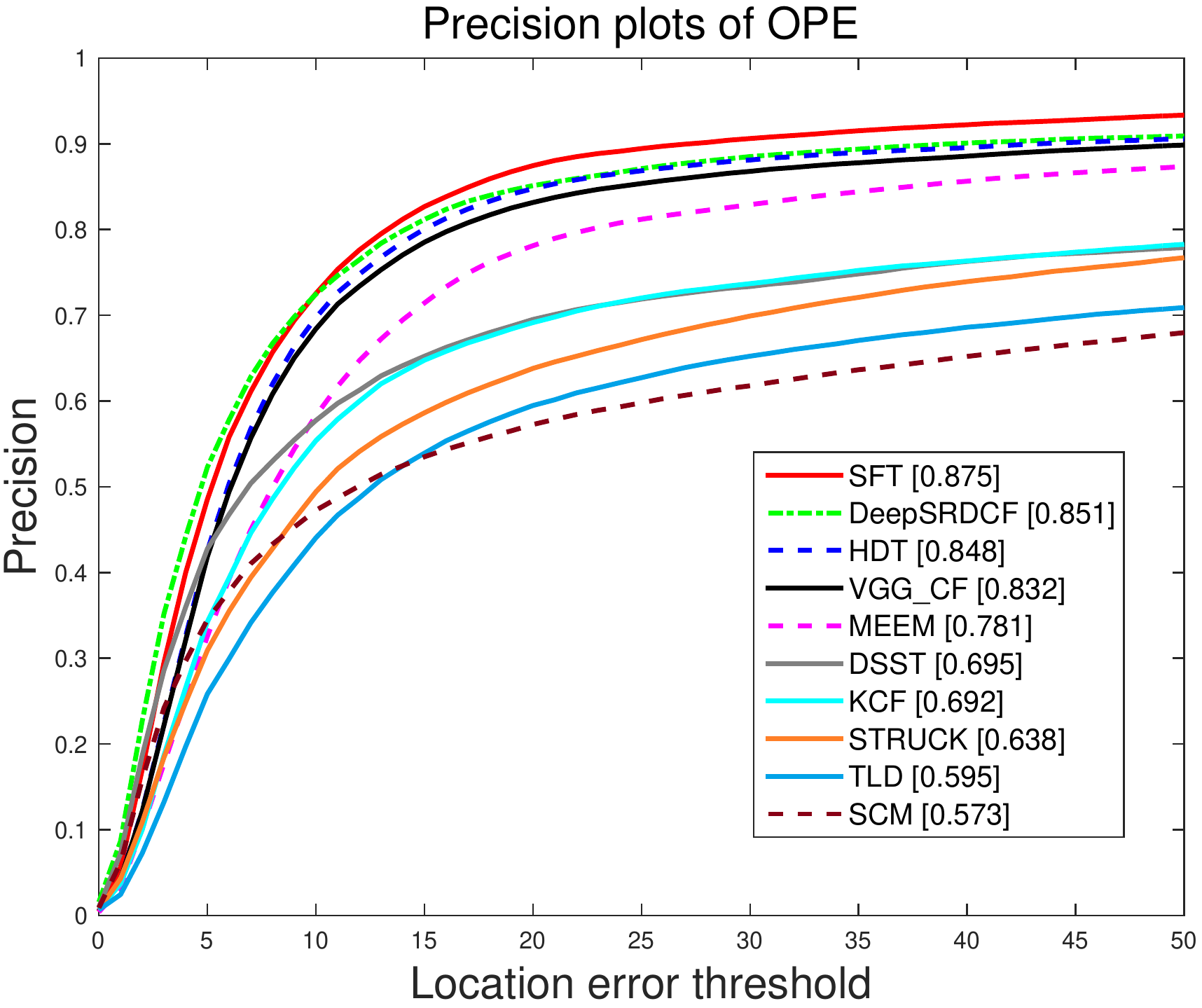}}
\subfigure[]{\label{fig:6-2}
\includegraphics[width=0.47\linewidth]{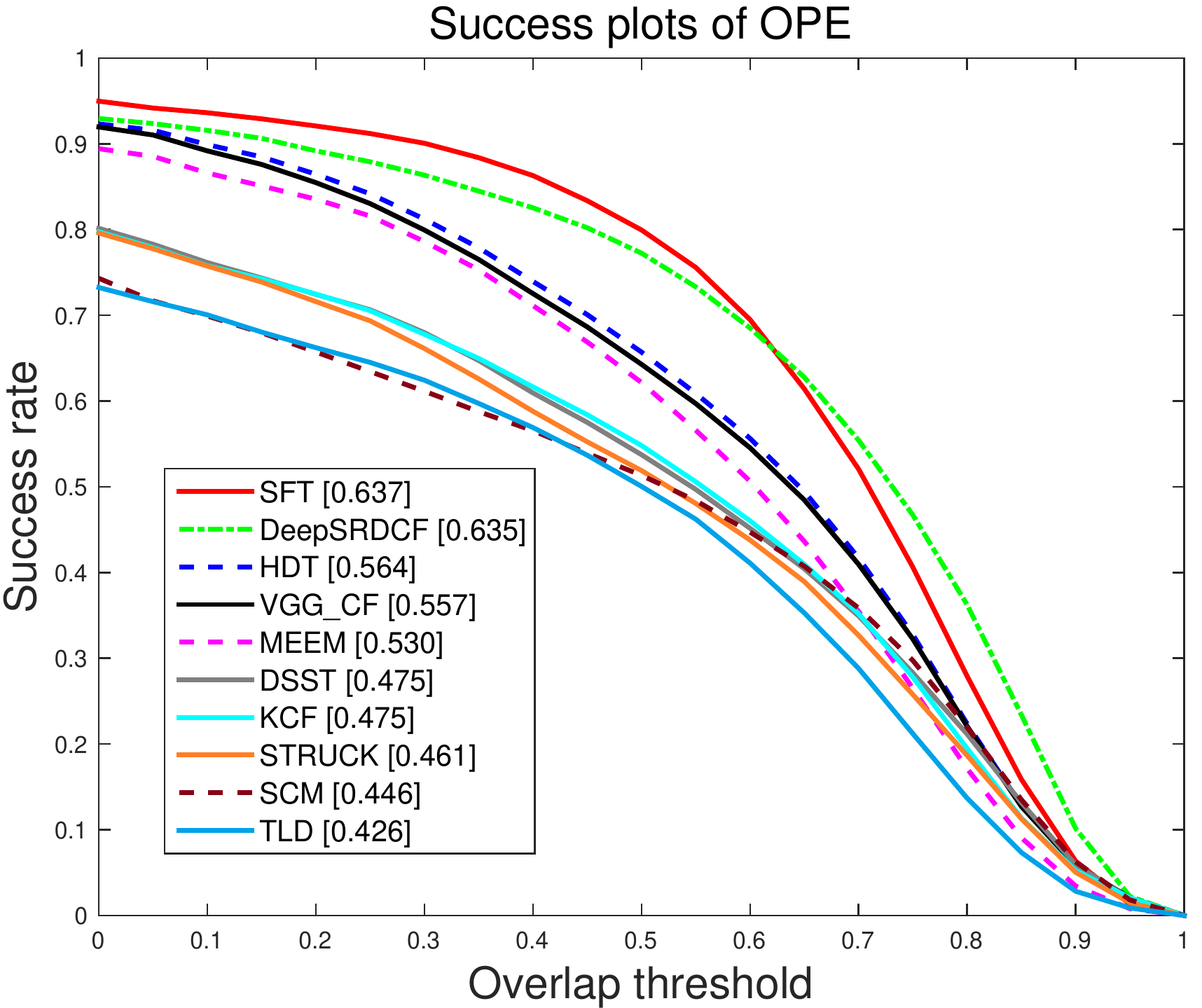}}
\caption{The precision and success plots of quantitative comparison for the 100 sequences on OTB-2015 dataset. The center location error (CLE) and area under curve (AUC) scores of the top 10 trackers are reported. Best viewed with Zooming up.}
\label{fig:compare-stateofart}
\end{figure}

\textbf{Quantitative Evaluation.} Fig.~\ref{fig:compare-stateofart} plots the precision curves and success curves among all trackers. The top-10 trackers ranked by CLE and AUC scores are shown with different colors. From these figures, we have three observations: (i) In the precision plot, SFT outperforms all state-of-the-art trackers, which demonstrates its effectiveness. (ii) In success plot, SFT achieves a comparable result with the current best method DeepSDCF. Actually here we only use the parameters used in~\cite{danelljan2014accurate} without any tuning for scale estimation. (iii) In fine localization, SFT is slightly inferior to DeepSRDCF, but SFT is more robust to those large appearance variations. The main reason is that, the spectral filtering averages all features of neighbors when all edges are assigned to equal weights, thus in the filtering some subtle textures may be lost while most invariant information is preserved.


\begin{figure*}[!ht]
\centering
\subfigure{\label{fig:12-01}
\includegraphics[width=0.24\linewidth]{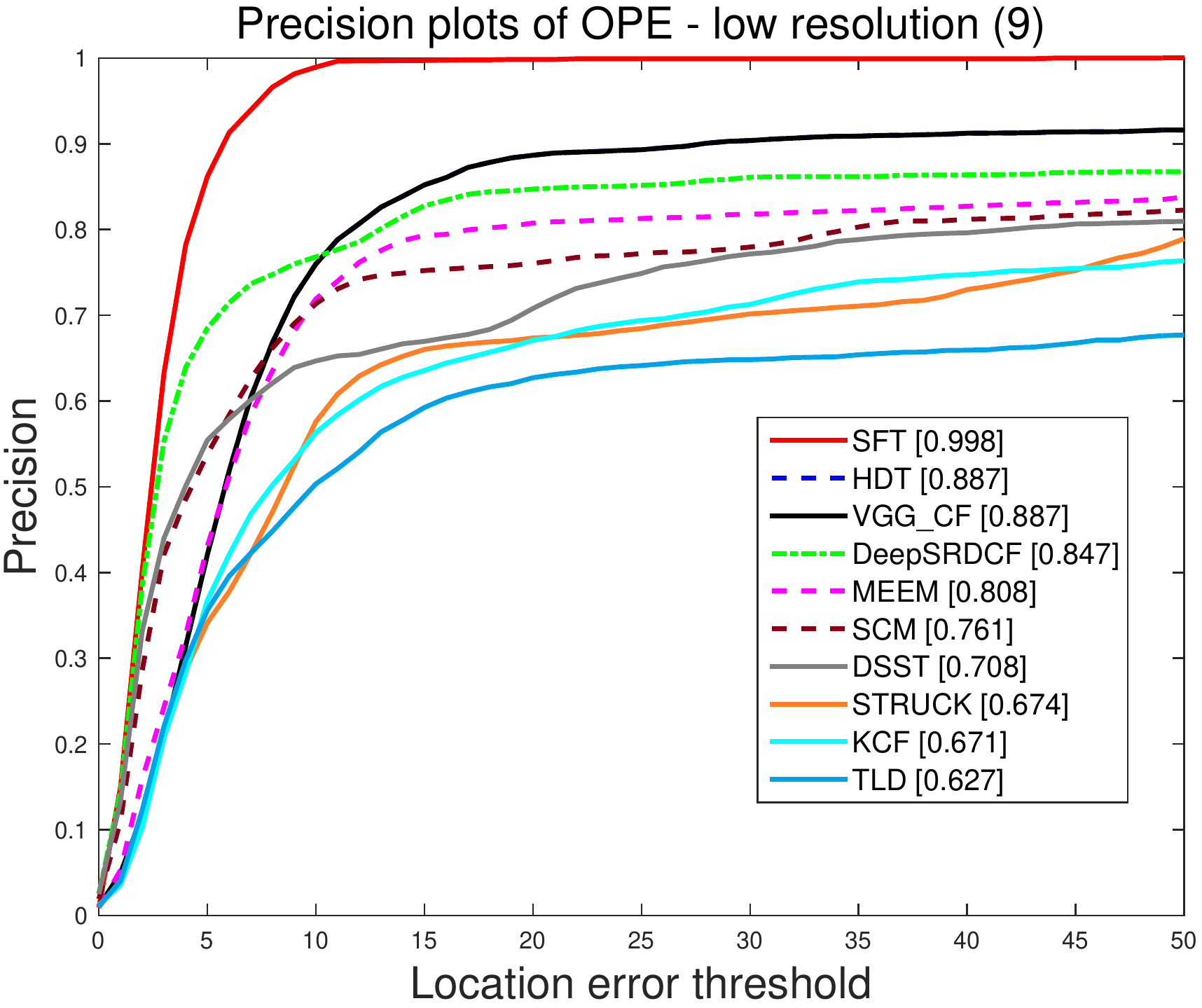}}
\subfigure{\label{fig:12-02}
\includegraphics[width=0.24\linewidth]{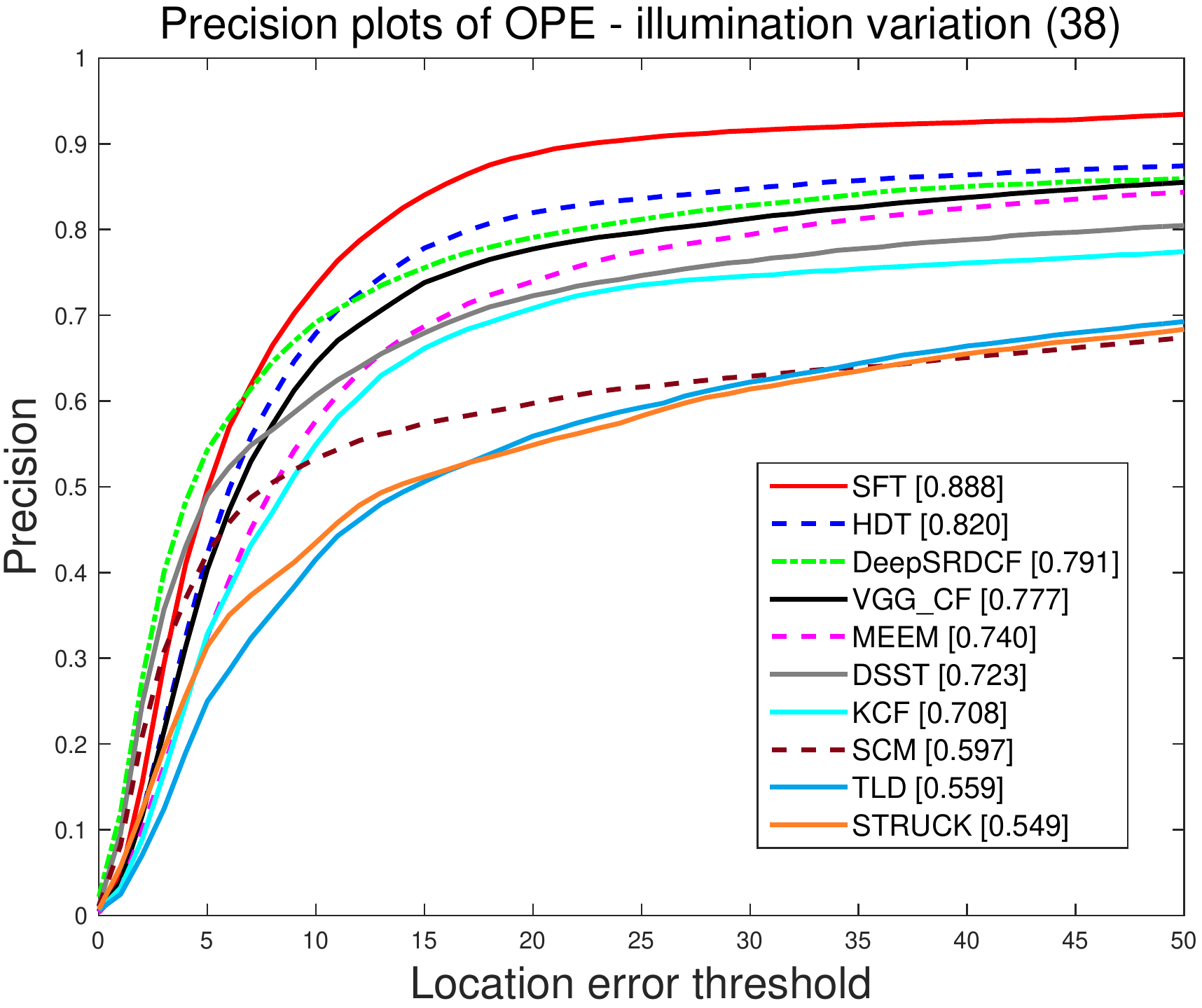}}
\subfigure{\label{fig:12-03}
\includegraphics[width=0.24\linewidth]{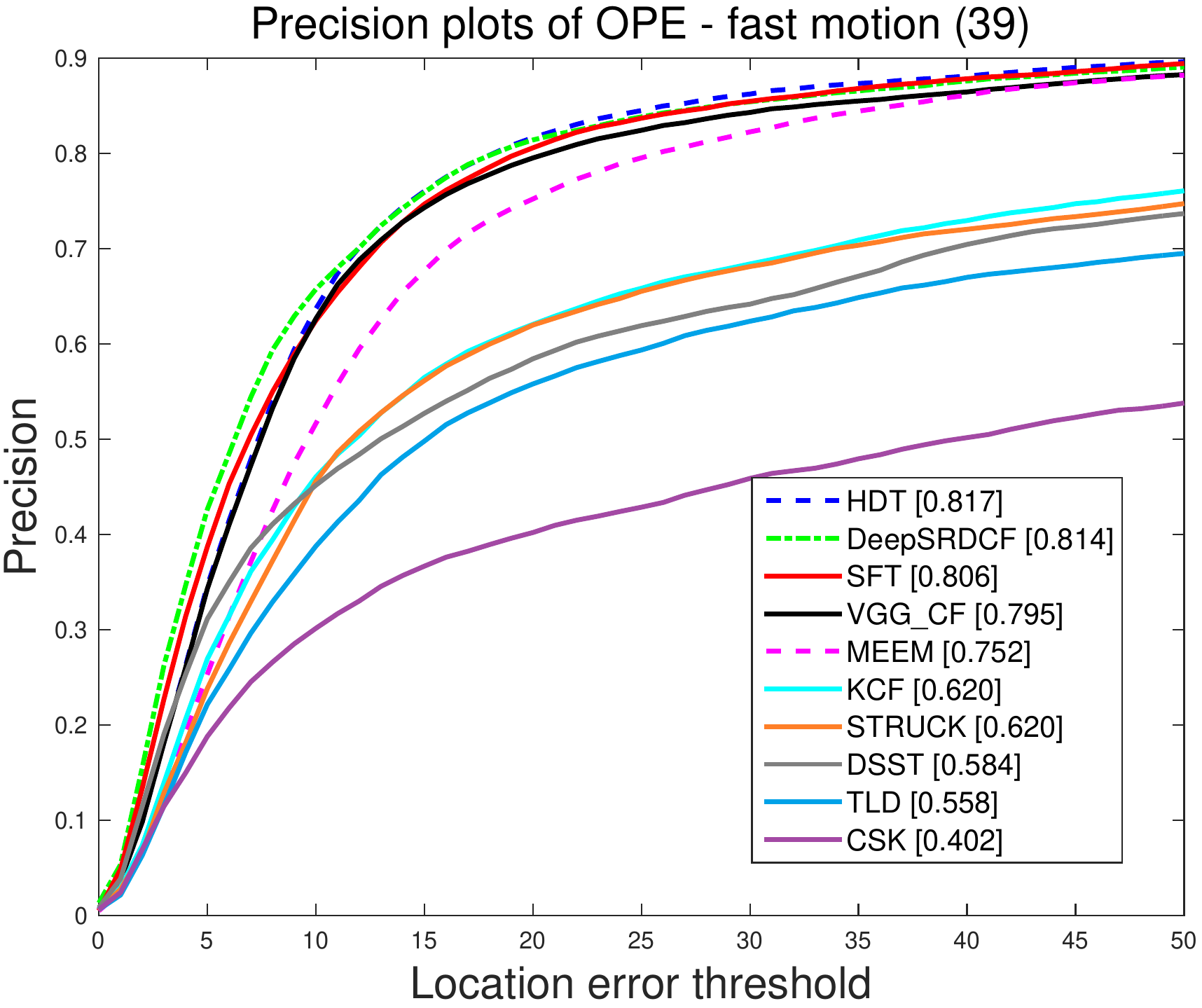}}
\subfigure{\label{fig:12-04}
\includegraphics[width=0.24\linewidth]{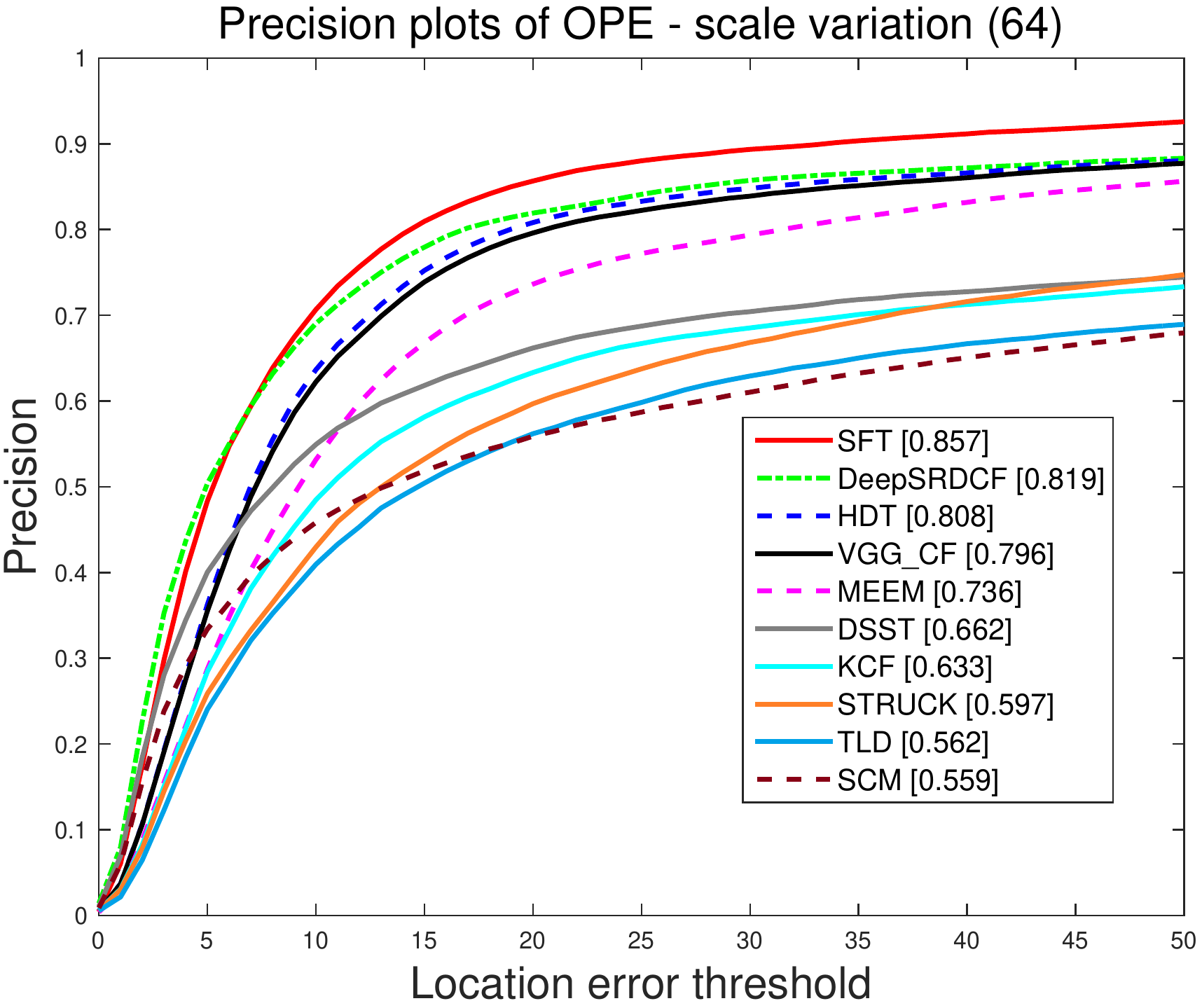}}
\subfigure{\label{fig:12-05}
\includegraphics[width=0.24\linewidth]{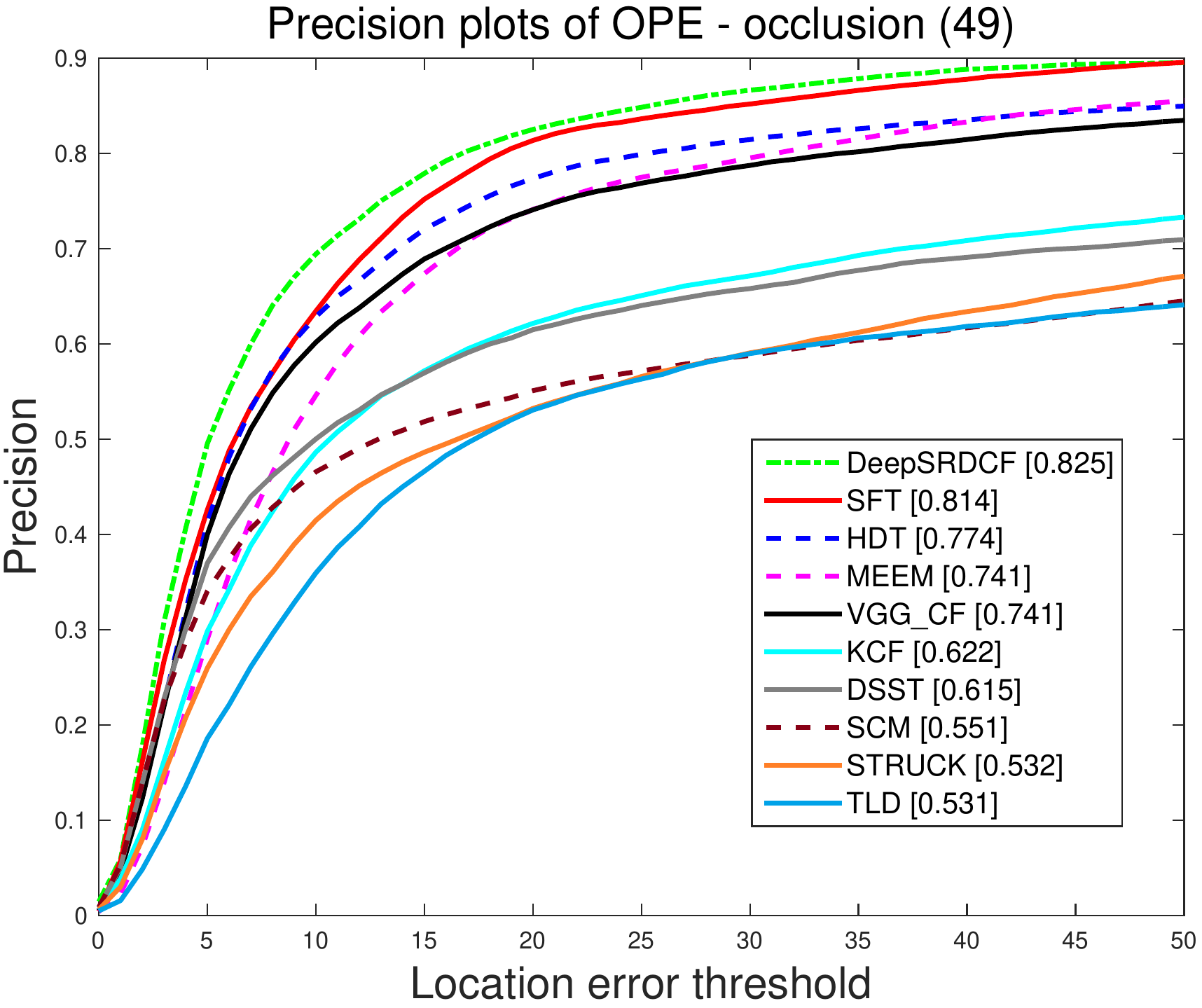}}
\subfigure{\label{fig:12-06}
\includegraphics[width=0.24\linewidth]{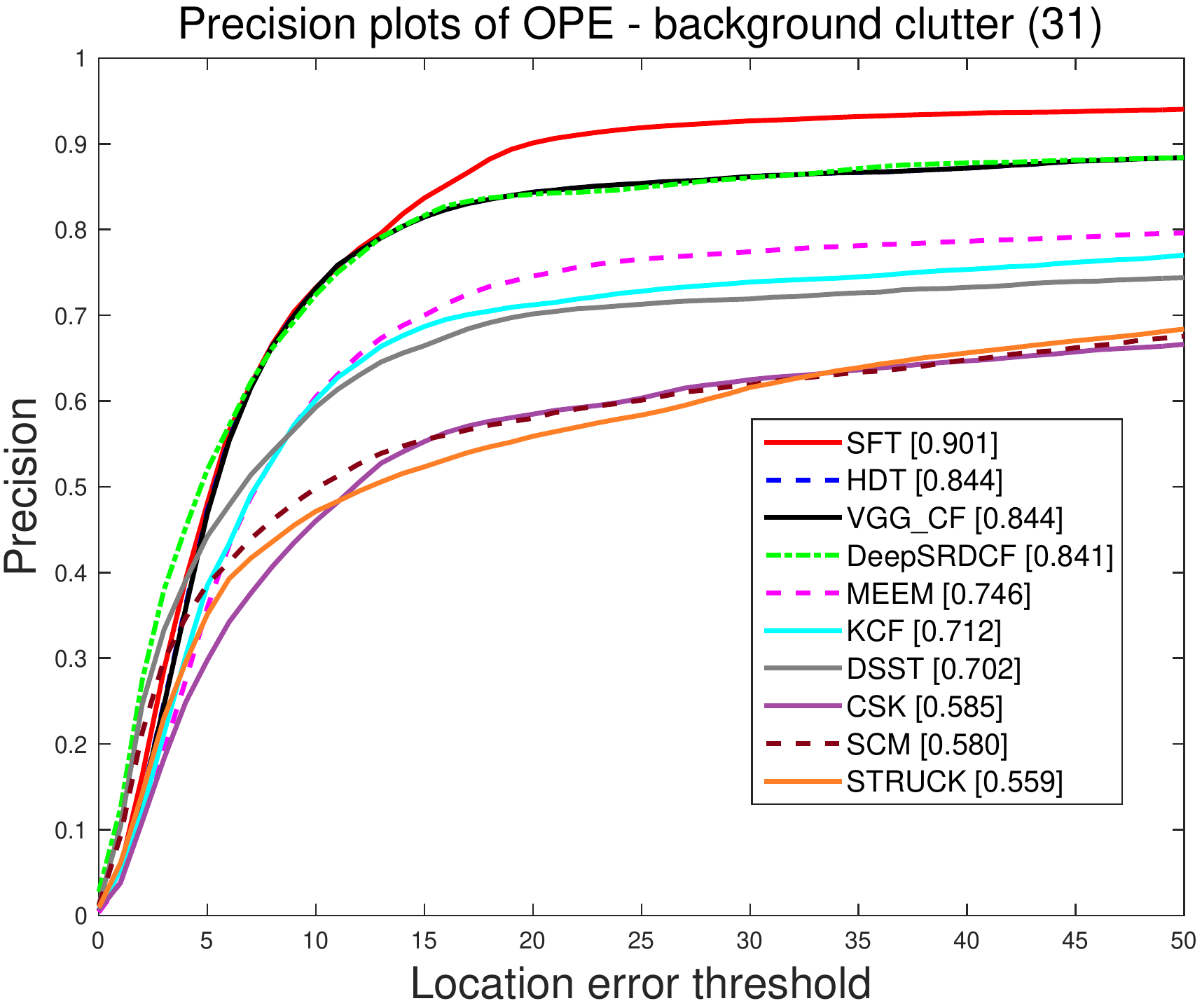}}
\subfigure{\label{fig:12-07}
\includegraphics[width=0.24\linewidth]{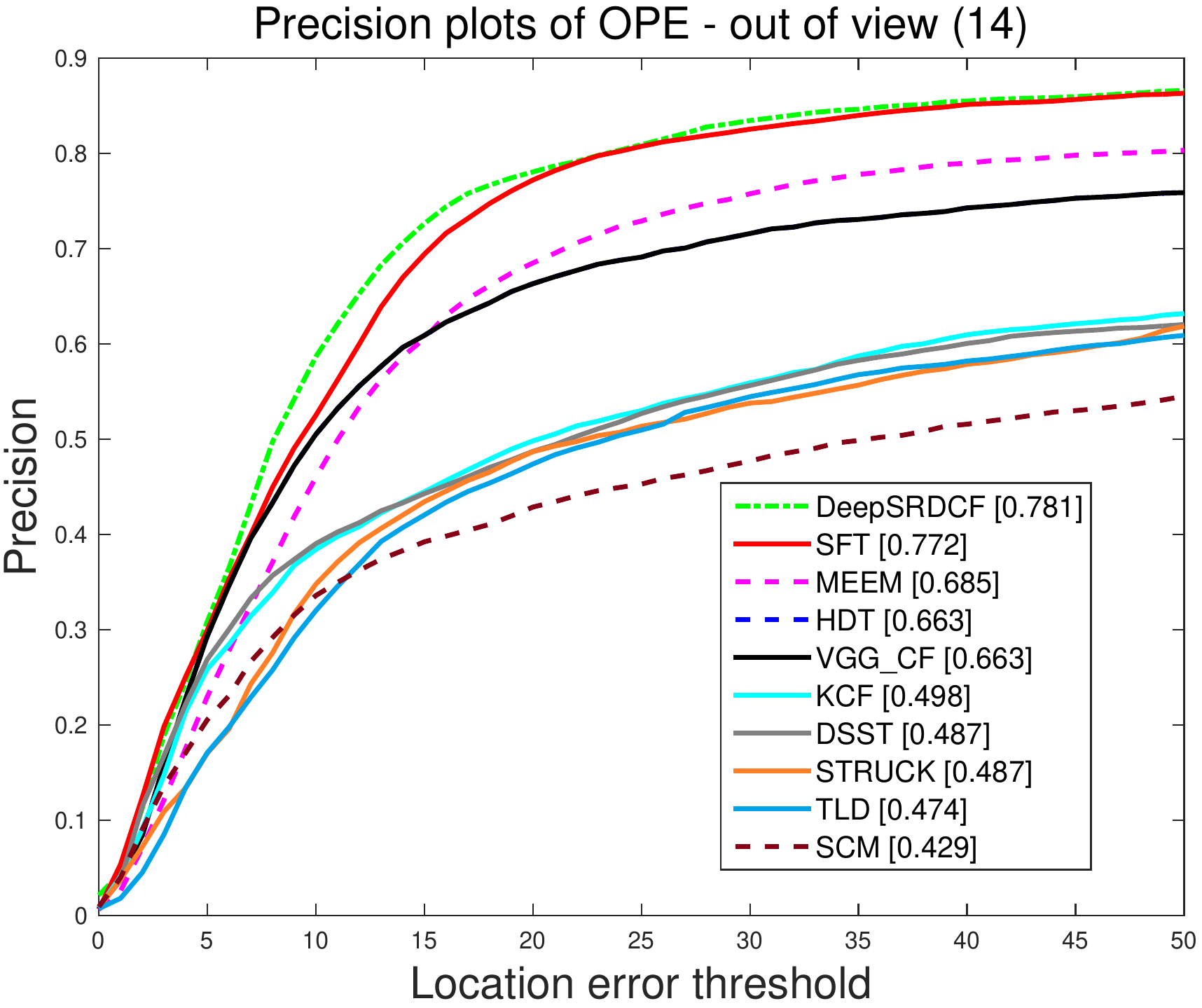}}
\subfigure{\label{fig:12-08}
\includegraphics[width=0.24\linewidth]{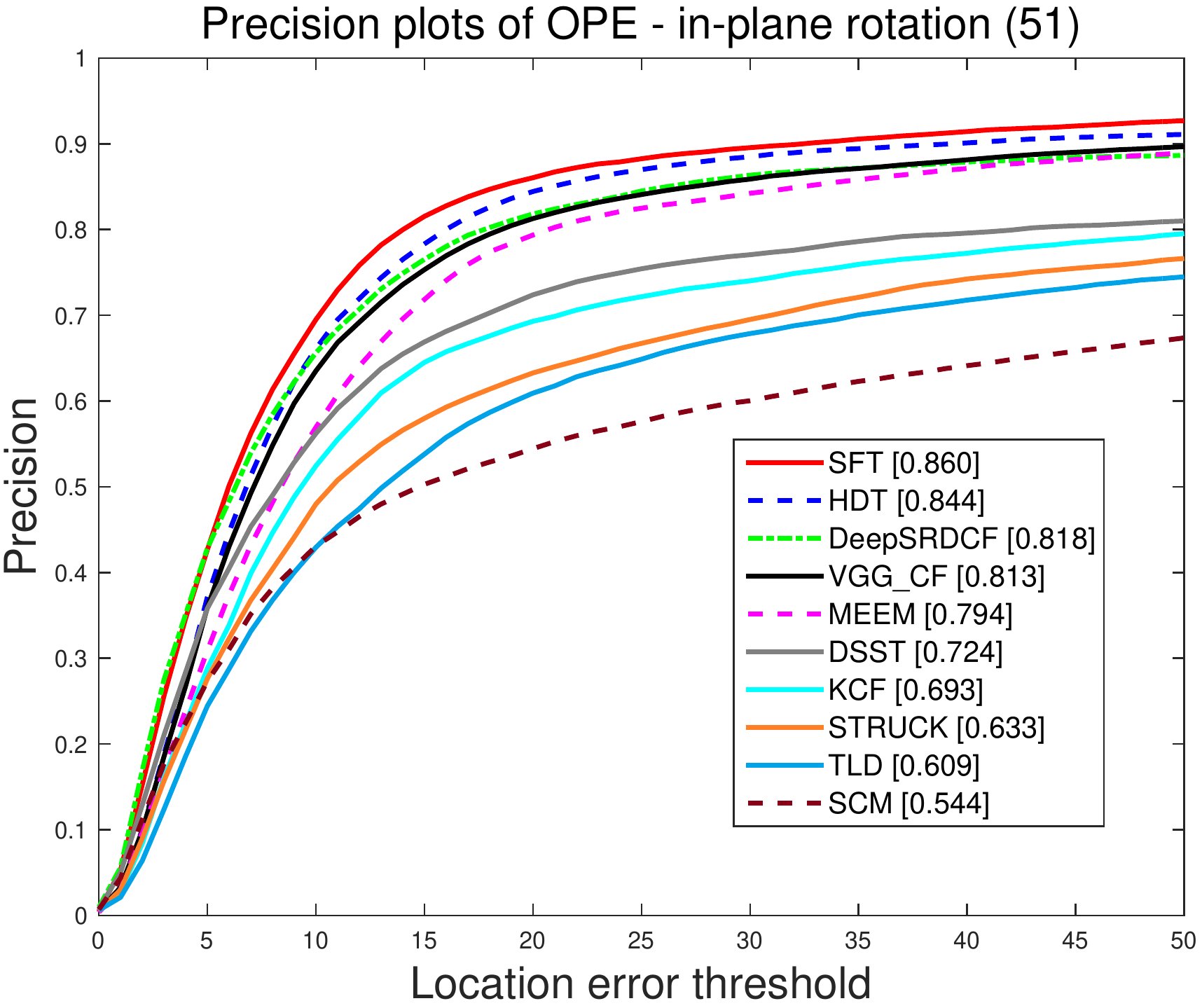}}
\subfigure{\label{fig:12-09}
\includegraphics[width=0.24\linewidth]{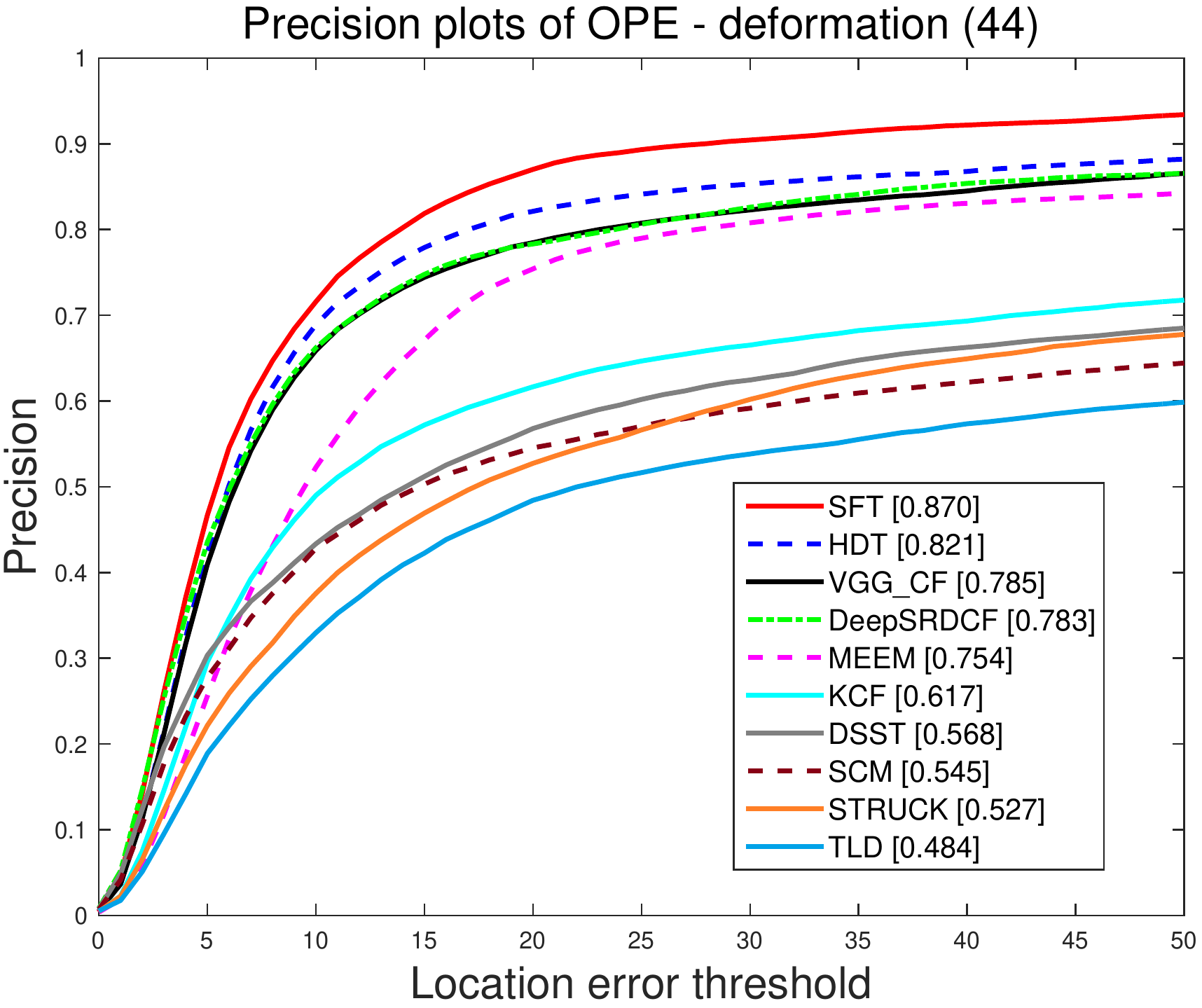}}
\subfigure{\label{fig:12-10}
\includegraphics[width=0.24\linewidth]{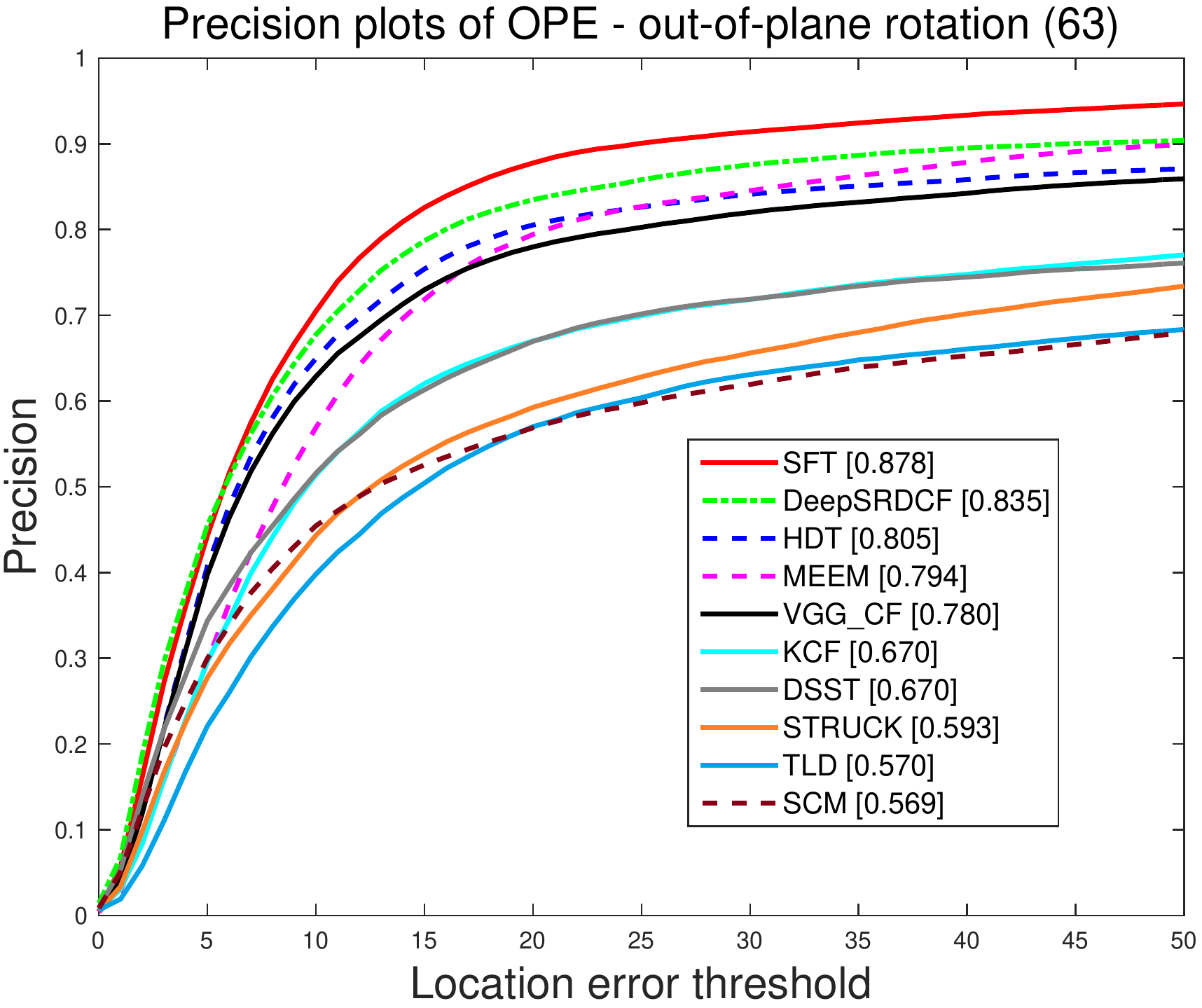}}
\subfigure{\label{fig:12-11}
\includegraphics[width=0.24\linewidth]{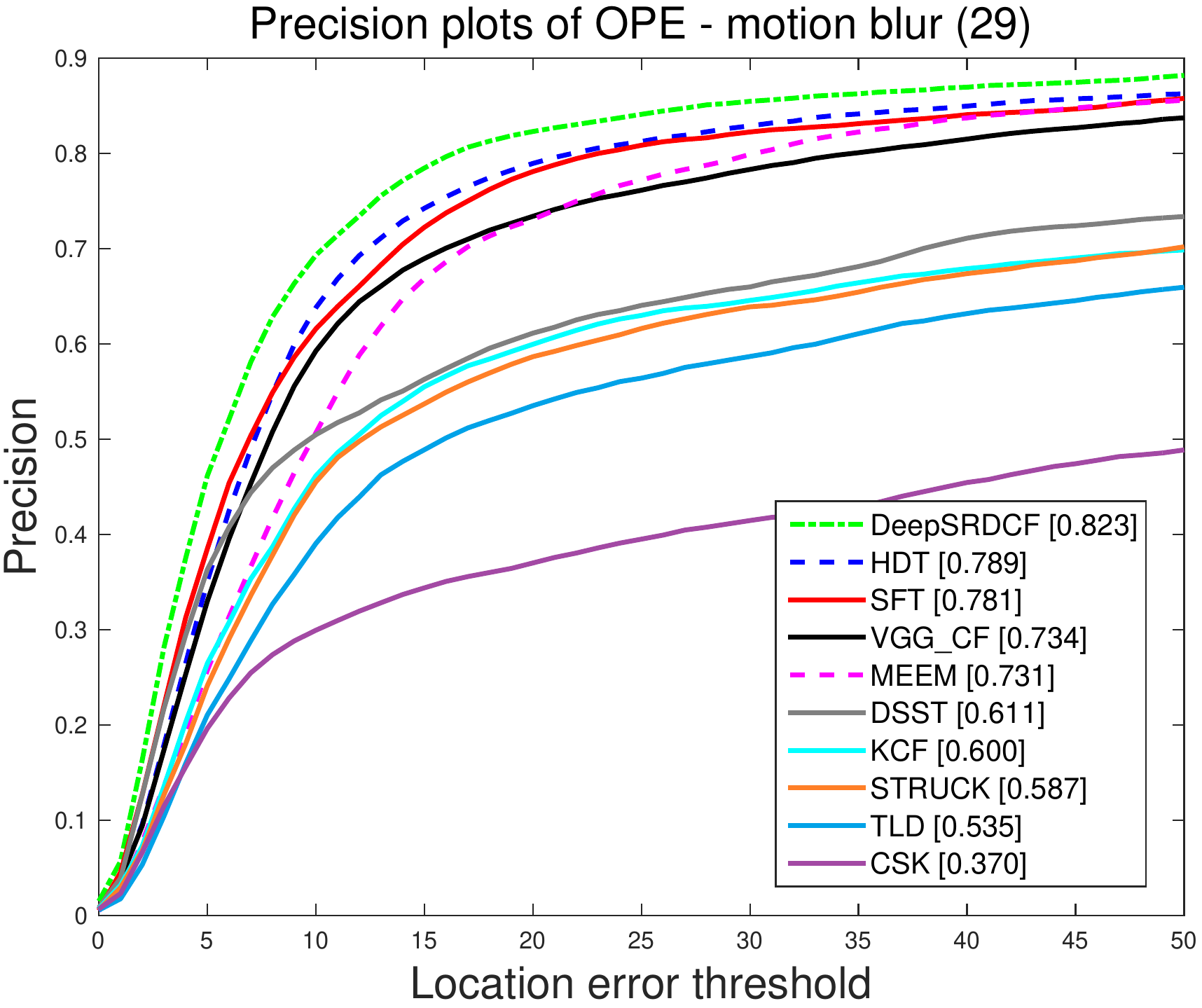}}
\subfigure{\label{fig:12-12}
\includegraphics[width=0.24\linewidth]{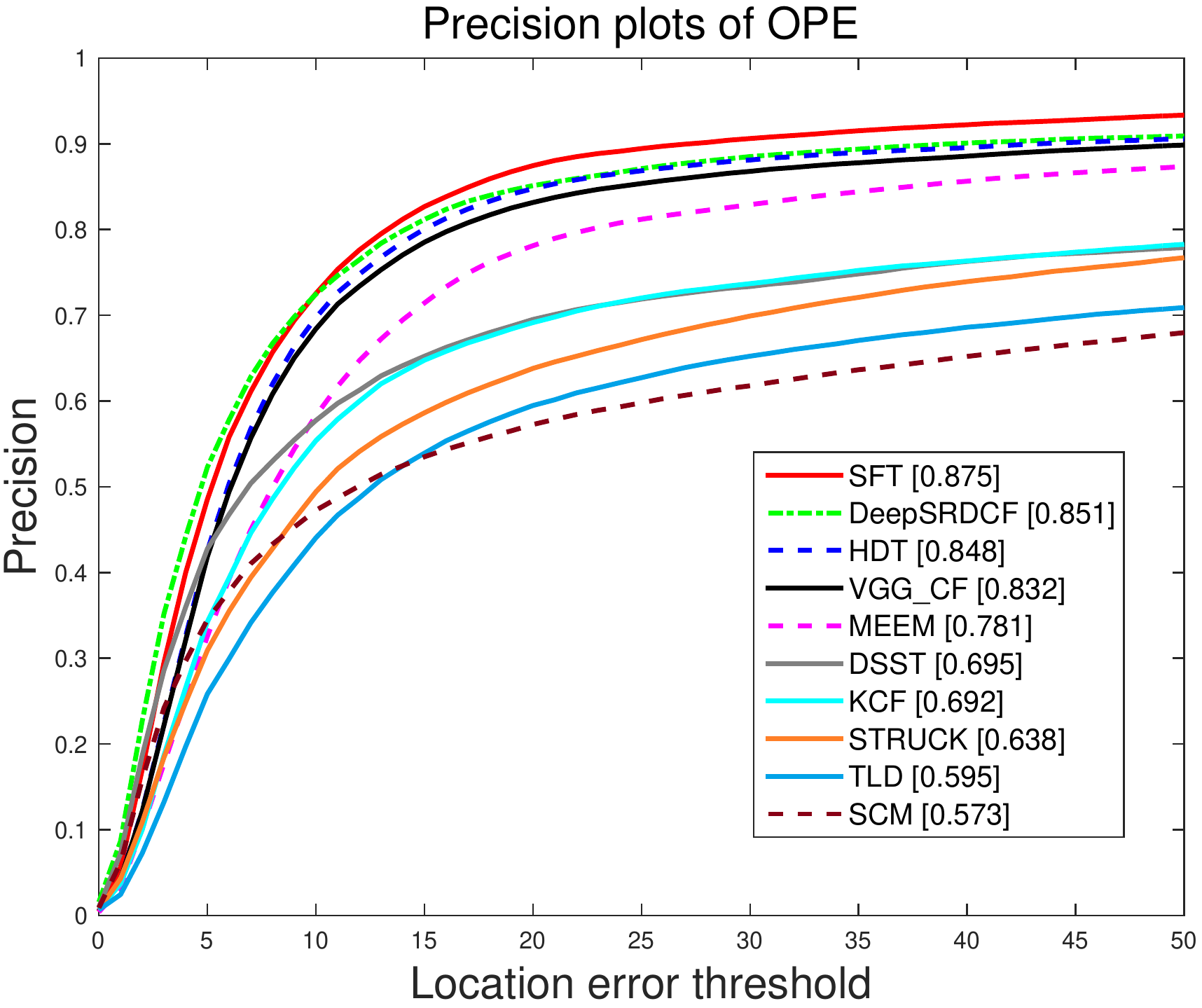}}
\caption{The precisions of plots of 11 attributes. Our tracker achieves is superior to other methods in most cases except motion blur and fast motion. The reasons of failures might be the small search window (2.4$\times$) and boundary effects for our method. Best viewed with Zooming up.}
\label{fig:12figs}
\end{figure*}

\begin{figure*}[!ht]
\centering
\includegraphics[width=0.98\linewidth]{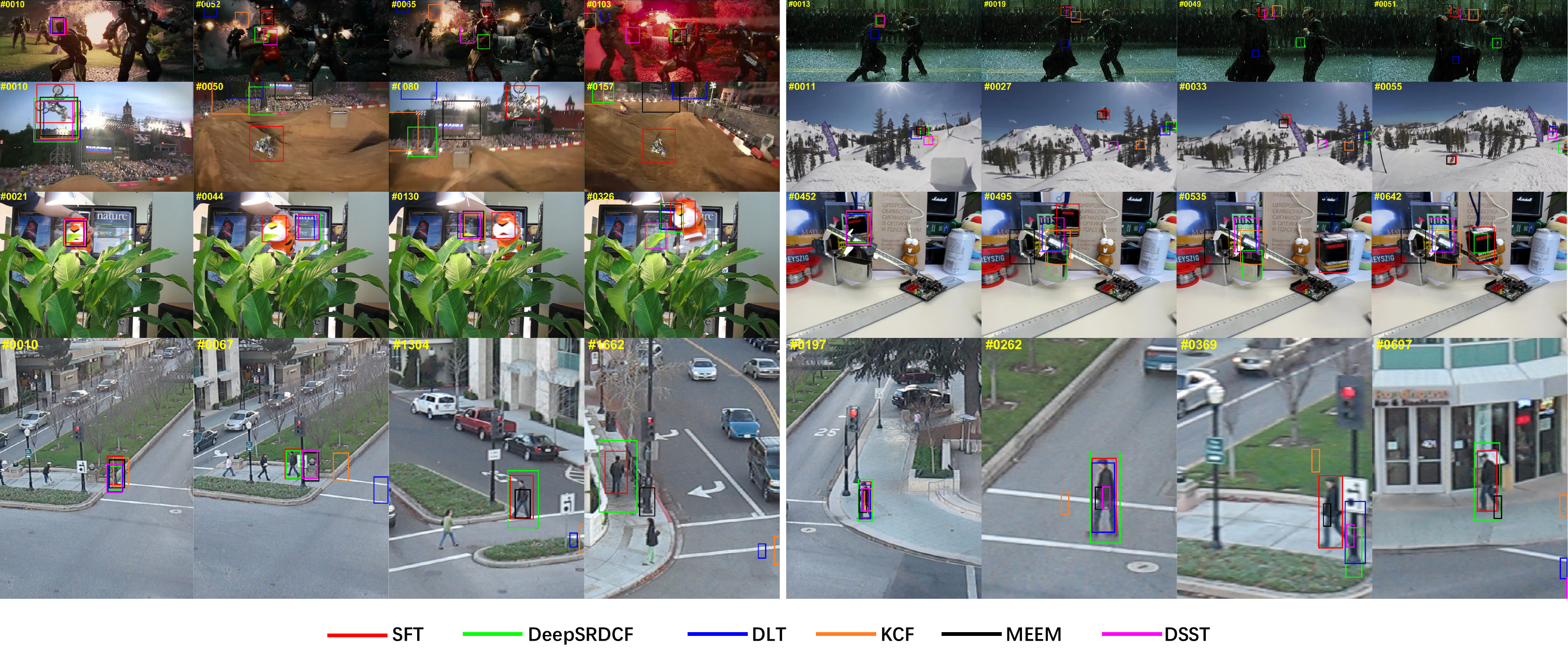}
\caption{Qualitative results of SFT on some most challenging sequences (Ironman, Matrix, MotorRolling, Skiing, Tiger1, Box, Human3, Human6 respectively from left to right and top to bottom). Best viewed with Zooming up.}
\label{fig:QualitativeResults}
\end{figure*}

\textbf{Attribute-based evaluation.} For comprehensive analysis of our proposed SFT, we provide each attribute plot in Fig.~\ref{fig:12figs}. As observed from these figures, SFT achieves more excellent performance compared to state-of-the-art in almost all cases. Particularly, SFT is rather effective in handling low resolution, background cluster and illumination variation challenges. In the case of low resolution the CLE score of our tracker is 99.8\% which surpasses DeepSRDCF by 11.1\%.  However, our tracker seems lost target easily in the case of fast motion and motion blur, which might be attribute to the small search window of SFT or boundary effect.

\textbf{Qualitative evaluation.} Fig.~\ref{fig:QualitativeResults} shows some visual results of the top ranked trackers (including our proposed SFT, DeepSRDCF, KCF, MEEM, DSST, DLT~\cite{wang2013learning}) on the eight most challenging image sequences, Ironman, Matrix, MotorRolling, Skiing, Tiger1, Box, Human3, Human6. AS shown in Fig~\ref{fig:QualitativeResults}, the prediction position and bounding box of our proposed method are more precise than others trackers in various scenes.

\section{Conclusion}

In this paper, we propose a simple but efficient Spectral Filter Tracking (SFT) method. In SFT, the candidate image region is model as a pixel-level grid graph. To estimate the best-matching vertex, we borrow spectral graph filters to encode the local graph structure. Considering the computation cost of eigenvalue decomposition on the Laplacian matrix, we approximate spectral filtering as the polynomial of a series of filter bases. For the filter bases, we employ the Chebyshev expansion terms, where each term encodes a localized filtering region of graph. Thus all filter bases span a multi-scale filtering space. Finally the filtering parameters and feature projecting function are jointly reduced into a simple regression model. The proposed SFT simply boils down to only a few line codes, but the experimental results on the dataset~\cite{wu2015object} demonstrate that the SFT is more effective and achieves state-of-the-art performance. In future, we will consider to speed up the tracker.


%
%
%


\end{document}